\documentclass[a4paper,fleqn]{cas-dc}

\usepackage[numbers]{natbib}

\usepackage{colortbl}
\usepackage{xcolor}

\usepackage{algorithm}
\usepackage{algpseudocode}
\algblock{Input}{EndInput}
\algnotext{EndInput}
\algblock{Output}{EndOutput}
\algnotext{EndOutput}

\usepackage{float} 
\usepackage{tabularx}
\usepackage{ragged2e} 
\usepackage{xcolor}
\usepackage{mdframed}
\usepackage{tikz}
\usepackage{subcaption}

\definecolor{lowcolor}{RGB}{255, 0, 0}

\definecolor{mediumcolor}{RGB}{255,165,0}

\definecolor{highcolor}{RGB}{0, 128, 0}

\definecolor{lightgray}{RGB}{211, 211, 211}

\usepackage{soul}
\usepackage{color}
\usepackage{caption}
\usepackage{colortbl}

\usepackage{xparse}

\ExplSyntaxOn
\NewDocumentCommand{\avercalc}{O{4}m}{
  \clist_set:Nn \l_tmpa_clist {#2}
  \fp_zero:N \l_tmpa_fp 
  \clist_map_inline:Nn \l_tmpa_clist {
    \fp_add:Nn \l_tmpa_fp {##1}
  }
  \fp_eval:n { round(\l_tmpa_fp / \clist_count:N \l_tmpa_clist, #1) } 
}
\ExplSyntaxOff

\def\tsc#1{\csdef{#1}{\textsc{\lowercase{#1}}\xspace}}
\tsc{WGM}
\tsc{QE}
\tsc{EP}
\tsc{PMS}
\tsc{BEC}
\tsc{DE}

\begin{document}
\let\WriteBookmarks\relax
\def\floatpagepagefraction{1}
\def\textpagefraction{.001}
\shorttitle{The Performance of the LSTM-based Code Generated by Large Language Models....}
\shortauthors{Saroj Gopali et~al.}

\title [mode = title]{The Performance of the LSTM-based Code Generated by Large Language Models (LLMs) in Forecasting Time Series Data}                      




\author[1]{Saroj Gopali}[orcid=0000-0003-3565-9756]
\cormark[1]
\ead{saroj.gopali@ttu.edu}


\affiliation[1]{organization={Department of Computer Science, Texas Tech University},
                country={USA}}

\author[2]{Sima Siami-Namini}[orcid=0000-0002-3758-4172]
\ead{ssiamin1@jh.edu}
\affiliation[2]{organization={Advanced Academic Programs, Johns Hopkins University},
                country={USA}}

\author[3]{Faranak Abri}[orcid=0000-0003-3028-094X]
\ead{faranak.abri@sjsu.edu}
\affiliation[3]{organization={Department of Computer Science, San Jose State University},
                country={USA}}

\author[1]{Akbar Siami Namin}[orcid=0000-0002-1646-7495]
\ead{akbar.namin@ttu.edu}

\cortext[1]{Corresponding author: Department of Computer Science, Texas Tech University}
\cortext[cor2]{Principal corresponding author}


\begin{abstract}
Generative AI, and in particular Large Language Models (LLMs), have gained substantial momentum due to their wide applications in various disciplines. While the use of these game changing technologies in generating textual information has already been demonstrated in several application domains, their abilities in generating complex models and executable codes need to be explored. As an intriguing case is the goodness of the machine and deep learning models generated by these LLMs in conducting automated scientific data analysis, where a data analyst may not have enough expertise in manually coding and optimizing complex deep learning models and codes and thus may opt to leverage LLMs to generate the required models. This paper investigates and compares the performance of the mainstream LLMs, such as ChatGPT, PaLM, LLama, and Falcon, in generating deep learning models for analyzing time series data, an important and popular data type with its prevalent applications in many application domains including financial and stock market. 
This research conducts a set of controlled experiments where the prompts for generating deep learning-based models are controlled with respect to sensitivity levels of four criteria including 1) Clarify and Specificity, 2) Objective and Intent, 3) Contextual Information, and 4) Format and Style. While the results are relatively mix, we observe some distinct patterns. We notice that using LLMs, we are able to generate deep learning-based models with executable codes for each dataset seperatly whose performance are comparable with the manually crafted and optimized LSTM models for predicting the whole time series dataset. We also noticed that ChatGPT outperforms the other LLMs in generating more accurate models. Furthermore, we observed that the goodness of the generated models vary with respect to the ``temperature'' parameter used in configuring LLMS. The results can be beneficial for data analysts and practitioners who would like to leverage generative AIs to produce good prediction models with acceptable goodness.

\end{abstract}




\maketitle

\section{Introduction}
\label{sec:intro}

Large language models (LLMs) such as ChatGPT \cite{ouyang2022training}, LLaMa \cite{Llama1}, Falcon \cite{penedo2023refinedweb}, and PaLM \cite{chowdhery2022palm} are gaining popularity on a regular basis and for a variety of reasons. These generative models are already playing an integral role in assisting people with their day-to-day duties, such as generating code, writing emails, assisting with projects, and many more. As a result, a wider range of users is involved in dealing with LLMs. 
According to a report published by Markets and Markets Research Pvt. Ltd.\cite{MarketsandMarkets}, the global market for generative AI is expected to achieve at a Compound Annual Growth Rate (CAGR) of 35.6$\%$ between 2023 and 2028, indicating significant potential opportunities. The CAGR was valued at \$11.3 billion in 2023 and it is expected to rise to almost \$51.8 billion by 2028. Looking ahead, industry forecasts estimate that the value of generative AI market might reach \$191.8 billion by 2032.

Recent research suggests that generative AI-based applications could contribute an annual value ranging from $\$2.6$ trillion to $\$4.4$ trillion across various use cases, surpassing the 2021 GDP of the United Kingdom at $\$3.1$ trillion \cite{Zemmel_2023}. Such integration could enhance the overall influence of artificial intelligence by up to 40$\%$ with the possibility of further doubling this estimate by incorporating generative AI into existing software applications used for tasks beyond the initially analyzed use cases. 

As per Salesforce's findings \cite{Goldman_2023}, $61\%$ of employees already employ or intend to utilize generative AI for accomplishing their tasks. Additionally, $68\%$ of employees believe that generative AI can enhance their ability to serve customers better. Moreover, $67\%$ of employees feel that it can amplify the benefits derived from other technological investments. These insights highlight the growing adoption of LLMs in various professional settings. 

In this research work, we have studied the performance of the following large language models:
\begin{itemize}
\item GPT$-$3.5$-$Turbo \footnote{https://platform.openai.com/docs/models/gpt-3-5} is an efficient model that are designed for natural languages and code comprehension. 
\item Falcon \cite{penedo2023refinedweb}, is one of the most effective and optimized language models based on high quality, large scale training data. 
\item Meta AI’s Llama 2\cite{touvron2023llama}, is a series of large language models with 7 to 70 billion parameters and it is excellent in knowledge, reasoning, and code benchmarks.
\item Google’s PaLM \cite{chowdhery2022palm}, a 540B parameter model is highly proficient across various applications such as translation, QA pipelines, and even arithmetic.
\end{itemize}
These Large Language Models have been leveraged to perform analysis and model building tasks automatically, by providing prompts in a Natural Language processing format. LLM models such as GPT$-$3.5$-$Turbo,  Falcon ,  LLama 2  and PaLM  are extensively integrated for the task in code generation \cite{ni2023lever}, text generation \cite{wang2023chatgpt} and image generation \cite{qu2023layoutllm} where the instructions are provided thought prompts in texts.

An interesting question is whether LLMs can also be leveraged by professional data analysts with expertise in certain domains (e.g., financial market) to help them generate a relatively good model (e.g., Long Short-Term Models - LSTM) and the corresponding executable code (e.g., Python) automatically without any additional needs for learning complex syntax and semantic of developing these deep learning-based forecasting models (e.g., LSTM) from scratch? In the context of anomaly detection on time series \cite{9671488} \cite{electronics11193205}, URL detection \cite{10.1145/3605098.3636164} and vulnerability detection in smart contracts \cite{9842527} LSTM model demonstrates exception performance, which is the primary reason of choosing the LSTM model. This paper conducts an exploratory analysis to investigate the performance of generative AIs, and in particular LLMs, and assess the goodness of the deep-learning codes generated by LLMs for building deep learning-based models in forecasting time series data. The underlying motivation is that most data analysts, who deal with time series data types, may need to design and develop their own complex deep learning-based codes.However, individuals not familiar with complex deep learning models often have limited knowledge to deal with building and training the deep learning model. They can generate code to build and train such models by prompting these large language models. The prompting approach makes deep learning more accessible for individuals who might have little or no experience but can take advantage of deep learning to work on their time series data.

The paper explores the prompts with controlled sensitive analysis based on categorical levels to study the goodness of models and codes generated by LLMs for deep learning-based time series analysis. The goal is to comprehensively evaluate and comprehend the influence of various category levels defined for prompts. Each prompt for generating time series analysis deep learning-based code is crafted based on the criteria including 1) Clarity and Specificity, 2) Objective and Intent, 3) Contextual Information, and 4) Format and Style. Furthermore, to assess the impact of each criterion on the goodness of the models created, we consider three classes of intensity or sensitivity level as expressed in each prompt including 1) high, 2) medium, and 3) low intensity where intensity refers to the amount of information given to LLMs through prompts. This paper makes the following key contributions:

\begin{enumerate}
\item We conduct a number of experiments where the sensitivity levels (i.e., Low, Medium, and High) of four criteria including Clarity and Specificity, Objective and Intent, Contextual Information, and Format and Styles are controlled. 
\item We report that LLMs are capable of generating relatively good models that are comparable with manually coded and optimized models and codes. 
\item We report that amongst the LLMs studied, ChatGPT outperformed in most cases generating more accurate models for predicting time series data in the context of financial and stock data. 
\item The results also show that the performance of LLMs vary with respect to the temperature parameter in the configuration when generating deep learning-based prediction models. 

\item We also report that we did not observe a clear benefit of crafting more complex and detailed prompts in generating better and more accurate models. The results are mix where in some cases  models generated with simple prompts outperform models with more complex prompts. The results seem to be dependent on the setting of the temperature parameter. 
\end{enumerate}


The rest of this paper is structured into the following sections. In Section \ref{sec:relatedWork}, relevant research studies have been discussed. Section \ref{sec:background} contains the preliminary background related to LLM (i.e. GPT-3.5-Turbo, Falcon, LLama-2, PaLM). Section \ref{sec:resarch_questions}  outlines the research questions addressed in this work. Section \ref{sec:experiment} presents the experimental design, including the dataset, prompt framework, LLM configurations, and performance metrics used to evaluate the deep learning-based codes and models generated by LLMs for time series analysis. Our methodology is discussed in Section \ref{sec:expermintal_study}. 
Section \ref{sec:results} reports the results and discussion, obtained by each LLMs across the categories and levels. Section \ref{sec:limitations} presents the limitations of the paper and Section \ref{sec:conclusion} summarizes the conclusions of the work.

\vspace{-0.2cm}
\section{Related Work}
\label{sec:relatedWork}

In November 2022, OpenAI released ChatGPT \cite{ouyang2022training}. In February 2023, Meta released LLaMa \cite{Llama1} followed by Technology Innovation Institute (TII) introducing ``Falcon LLM'' \cite{penedo2023refinedweb}, a foundational Large Language Model (LLM) with 40 billion parameters that was introduced in March 2023. In May 2023, Google joined the race and announced PaLM \cite{chowdhery2022palm}. Moreover, Meta continued to release models, offering a set of models in July 2023 under the name Llama 2 \cite{touvron2023llama}, with parameter counts ranging from 7 billion to 70 billion. Since then major high tech companies continue improving their LLMs by adding additional features and capabilities. 

The idea of leveraging generative AIs in building executable codes and models has been discussed in several research papers. Vaithilingam et al. \cite{vaithilingam2022expectation} conducted a study with 24 volunteers to evaluate the usability of GitHub Copilot, a code generation tool that employs sophisticated language models. Participants in the research completed Python programming tasks using both Copilot and a controled condition that used VSCode's default IntelliSense functionality. The research sought to ascertain the influence of these tools on programming experience, error detection, problem-solving tactics, and barriers to their adoption. According to quantitative research by the authors, there was no significant difference in job completion times between Copilot and IntelliSense controlled groups. However, it was discovered that Copilot customers had more failures, which were mostly related to Copilot's incorrect advice. Despite this, the majority of participants (19 out of 24) preferred Copilot because of its ability to give a useful and informative starting point that eliminate the needs for frequent Web searches. However, several participants had difficulty comprehending and debugging the code generated by Copilot.

Destefanis et al. \cite{destefanis2023preliminary} studied and compared the performance of two AI models: GPT-3.5 and Bard, in generating code for Java functions. The Java functions and their descriptions were sourced from CodingBat Website, a platform for practicing programming problems. The evaluation of the Java code generated by the models was based on correctness, which was further verified using CodingBat's test cases. The results of the evaluation showed that GPT-3.5 outperformed Bard in code generation, producing accurate code for around $90.6\%$ of the functions, while Bard achieved correctness for only $53.1\%$ of the functions. Both AI models displayed strengths and weaknesses. GPT-3.5 consistently performed better across most problem categories, except for functional programming, where both models showed similar performance.

Liu et al. \cite{liu2023your} proposed EvalPlus, a framwork for rigorously evaluating the functional correctness of code generated by large language models (LLMs). The framework solved the issue of insufficient test coverage in current coding benchmarks like as \texttt{HUMANEVAL}, which employ only a few manually written test cases and consequently miss numerous problems in LLM-generated code. EvalPlus is built around an automated test input generator that combine LLM and mutation-based methods. It begins by using ChatGPT to generate high-quality seed inputs focusing at edge situations. The seeds are then changed using type-aware operators to produce a large number of new test cases quickly.  
The findings showed that inadequate benchmark testing could have a significant impact on claimed performance. EvalPlus also found flaws in 11$\%$ of the original \texttt{HUMANEVAL} solutions. Through automated testing, the study points in the direction of thoroughly analyzing and refining programming benchmarks for LLM-based code creation.

Ni et al. \cite{ni2023lever} proposed LEVER, a method for improving language-to-code generation by Code Language Models (LLMs) utilizing trained verifiers, as proposed in their work. They trained different verifier models based on plain language input, program code, and execution outcomes to determine the validity of created programs. LEVER was tested on four language-to-code datasets: Spider, WikiTableQuestions, GSM8k, and MBPP, in the fields of semantic parsing, table quality assurance, arithmetic reasoning, and basic Python programming. LEVER enhanced execution accuracy over strong baselines by $4.6-10.9\%$ when paired with Codex and achieved new state-of-the-art outcomes on all datasets. The relevance of execution results in verification became clear through ablation study, and the technique kept its strong performance even in circumstances with limited resources and without supervision. The findings showed that using benchmark datasets to train compact verifiers increased the performance of various LLMs in the field of language-to-code generation.

Denny et al. \cite{denny2023promptly} proposed ``Prompt Problems'', a unique educational idea aimed to educate students on how to create effective natural language prompts for large language models (LLMs) with the objective of generating executable codes. The authors created Promptly, a web-based application that allows students to iteratively tweak prompts based on test case output until the LLM produces accurate code. They used Promptly in classroom research with 54 beginning Python students and discovered that the tool teaches students to new programming structures and promotes computational thinking, despite the fact that some students were hesitant to utilize LLMs. The research looked at prompt duration and iteration counts, as well as student opinions based on open-ended feedback. Overall, the work presents preliminary evidence that quick Problems warrant more investigation as an approach to developing the growing ability of quick engineering.

Becker et al. \cite{becker2023programming} investigated the revolutionary impact of AI-driven code generation tools like OpenAI Codex, DeepMind AlphaCode, and Amazon CodeWhisperer. These tools possess the remarkable ability to translate natural language prompts into functional code, heralding a potential revolution in the realm of programming education. While admitting their potential, the authors argue for urgent talks within the computer science education community in order to overcome difficulties and properly utilize these technologies. The study provided an overview of important code generation models—Codex, AlphaCode, and CodeWhisperer—that were trained on massive public code repositories. These models excel at creating code in several programming languages and go beyond coding by providing features such as code explanations and language translation. From examples,  answers, and different problem-solving methodologies to scalable learning materials and an emphasis on higher-level topics, code-generating tools provide potential in education.  
The authors underline the need for educators proactively integrate these technologies, anticipating ethical concerns and a trend toward code analysis.

Zamfrescu-Pereira et al. \cite{10.1145/3544548.3581388} conducted a study whose findings shed some light on the difficulties that non-AI specialists have when attempting to provide effective prompts for large language models like GPT-3. These individuals frequently use a more impromptu and ad hoc approach rather than a systematic one, which is hampered by a tendency to overgeneralize from limited experiences and is based on human-human communication conventions. The authors developed BotDesigner, a no-code chatbot design tool for iterative fast development and assessment. This tool helps with a variety of tasks, including dialogue formulation, error detection, and fast alteration testing. Participants in a user research adjusted prompts and assessed modifications well, but with limited systematic testing and issues in prompt efficacy understanding. These difficulties originate from a tendency to overgeneralize and predict human-like behaviors. Through patterns and cause analysis, the study proposed potential for further training and tool development to encourage systematic testing, moderate expectations, and give assistance, while noting persistent uncertainty regarding generalizability and social bias consequences. This experiment highlights the difficulties that non-experts have in rapid engineering and suggests to opportunities for more accessible language model tools.

Zhou el al. \cite{zhou2022large} introduce a novel approach called the Automatic Prompt Engineer (APE) designed to facilitate the automatic generation and selection of effective natural language prompts. The primary goal is to guide large language models (LLMs) towards desired behaviors. APE tackles this challenge by framing prompt generation as a natural language program synthesis problem. It treats LLMs as black box computers capable of proposing and evaluating prompt candidates. The APE method leverages LLMs in three distinct roles: 1) as inference models for suggesting prompt candidates, 2) as scoring models to assess these candidates, and 3) as execution models to test the selected prompts. Prompt candidates are generated either directly through inference or recursively by creating variations of highly-rated prompts. The final selection of the most suitable prompt is determined by maximizing metrics such as execution accuracy on a separate validation set. Importantly, APE achieves these outcomes without the need for gradient access or fine-tuning, relying instead on a direct search within the discrete prompt space. In the experimental phase, APE was put to the test across a range of tasks. It successfully addressed 24 instruction induction tasks, exhibiting performance on par with or surpassing human capabilities across all of them. Additionally, APE demonstrated its effectiveness on a subset of 21 BIG-Bench tasks, outperforming human prompts in 17 out of 21 cases.

\vspace{-0.2cm}
\section{Large Language Models Studied}
\label{sec:background}
This paper compares the performance of four LLMs including GPT, Falcon, LLama-2, and PaLM. 
\subsection{GPT-3.5-Turbo}
GPT-3.5-Turbo is an OpenAI-developed variant of the Generative Pre-trained Transformer 3. GPT-3.5 models include a wide variety of capabilities including natural language and code comprehension and creation. GPT-3.5-Turbo is the standout model in this series known for its exceptional capabilities and low cost of ownership. The GPT-3.5 model, designed for chat interactions, boasts exceptional capabilities while being remarkably cost-effective, priced at only one-tenth of the cost of the \texttt{text-davinci-003} model.

\subsection{Falcon}
The Technology Innovation Institute located in Abu Dhabi created the Falcon LLM \cite{penedo2023refinedweb}, a significant advancement in AI language processing that has revolutionized its potential. Within the Falcon series, namely Falcon-40B and Falcon-7B, distinct versions, each possessing specific merits, contribute to making Falcon LLM an inventive and adaptable solution suitable for diverse uses.

Falcon's creation involved tailored tools and a distinctive data flow approach. This system extracts valuable Web information for customized training, differing from methods by NVIDIA, Microsoft, and HuggingFace. Focus on large-scale data quality was critical, recognizing LLMs' sensitivity to data excellence. Thus, an adept pipeline is built for rapid processing and quality content from Web sources. Falcon's architecture was meticulously optimized for efficiency. Coupled with high-caliber data, this enables Falcon to notably surpass GPT-3, utilizing fewer resources.

Falcon is a decoder-only model with 40 billion parameters trained with 1 trillion tokens. The training took two months and made use of 384 GPUs on AWS. After rigorous filtration and de-duplication of data from CommonCrawl, the model's pretraining dataset was generated using web crawls with roughly five trillion tokens. Falcon's capabilities were also expanded by incorporating certain sources such as academic papers and social media debates. The model's performance was then evaluated using open-source benchmarks such as EAI Harness, HELM, and BigBench.

\subsection{LLama-2}

Meta AI created lama 2 \cite{touvron2023llama}, a new family of pretrained and fine-tuned large language models (LLMs). Llama 2 has characteristics ranging from 7 billion to 70 billion arameters. The pre-trained models are designed for a wide range of natural language activities, whilst the fine-tuned versions known as Llama 2-Chat are designed for discourse. Llama 2 was pretrained on 2 trillion publically accessible tokens utilizing an improved transformer architecture with advantages like as extended context and grouped-query attention. On knowledge, reasoning, and code benchmarks, Llama 2 surpassed other open-source pretrained models such as Llama 1, Falcon, and MPT. Llama 2-Chat aligns the models to be helpful and safe in discourse by using supervised fine-tuning and reinforcement learning with human feedback (RLHF). Over 1 million fresh human preference annotations were collected in order to train and fine-tune reward models. To increase multi-turn discourse consistency, techniques such as Ghost Attention were created. Ghost Attention (GAtt) is a straightforward technique influenced by Context Distillation \cite{bai2022constitutional}. GAtt manipulates the fine-tuning stage to guide attention concentration through a step-by-step approach.

 \subsection {PaLM}

The Pathways Language Model (PaLM) \cite{chowdhery2022palm} is a Transformer based language model built by Google with 540 billion parameters. PaLM implements a traditional Transformer decoder structure with modifications such as SwiGLU activation and parallel layers for faster training. A total of 780 billion tokens of training data from natural language sources such as books, Web material, Wikipedia, GitHub code, and conversations were employed. The Pathways system, which allows for efficient distributed training on accelerators, was used to train on 6144 TPU v4 processors. This allows the training of such a big model without the need for pipeline parallelism.

The PaLM is evaluated over a wide range of tasks and datasets, proving its strong performance across several domains. The PaLM 540B achieved an outstnading score of 92.6 in the SuperGLUE test after fine tuning, essentially putting it with top models such as the T5-11B. In the field of question answering, PaLM 540B outperformed previous models by earning F1 scores of 81.4 on the Natural Questions and TriviaQA datasets in a few-shot setting. The model's abilities extended to mathematical thinking, where it achieved an astounding $58\%$ accuracy on the difficult GSM8K math word problem dataset using chain-of-thought cues. PaLM-Coder 540B has been elevated even further, reaching 88.4$\%$ success with a 
\texttt{pass@100} criterion on HumanEval and 80.8$\%$ success with a \texttt{pass@80} criterion on MBPP.

PaLM 540B excels at translation, earning a notable BLEU score of 38.4 in zero shot translation on the WMT English-French dataset, outperforming other significant language models. The model's responsible behavior is obvious in toxicity evaluations, with a RealToxicity dataset average toxicity probability of 0.46. Finally, using the WinoGrande coreference dataset, PaLM 540B achieved an accuracy of 85.1$\%$, demonstrating its capacity to mitigate gender bias. These extensive findings highlight the PaLM model's adaptability and efficacy across a wide range of language-related tasks.

\vspace{-0.2cm}
\section{Research Questions}
\label{sec:resarch_questions}

Recent advances in large language models like GPT-3 (Brown et al.\cite{brown2020language}) have demonstrated impressive text generation capabilities when provided with well-designed prompt instructions. However, best practices for prompt engineering are still developing. As Reynolds and McDonell  \cite{reynolds2021prompt} discuss, prompt programming,  the importance of being aware of prompts fitting within the concept of natural language.

This study will perform a systematic sensitivity analysis to identify the most sensitive prompt components for text generation using large language models. Following the workflow outlined by Saltelli et al. \cite{saltelli2008global}, each input factor will be varied individually while holding others constant to isolate its impacts. Text outputs will be analyzed to measure sensitivity.

Findings will provide prompt engineers with guidance on precision tuning. In line with recommendations by Pérez et al.  \cite{perez2021true}, this research aims to demystify prompts through empirical testing, allowing LM prompts and a few shots with hyperparameters.

\vspace{-0.2cm}
\section{Experimental Setup}
\label{sec:experiment}

The experiment was conducted in two parts. First, the LLM models were run in Google Colab Pro using a GPU with high RAM. Second, the outputs from the LLM models were run on a Macbook Pro with an M1 Max chip with 64 GB Memory.

\subsection{Dataset}

\begin{table}[h!]
    \centering
    \caption{Overview of Financial Dataset.}
    
    \scalebox{.72}{
    \begin{tabular}{|l|c|c|c|c|c|}
    \hline
    \multicolumn{1}{|c|}{\bf Dataset}  & \bf \# Data & \bf Start Date & \bf End Date  & \bf Sector &\bf Country\\ 
    
    \hline
        \bf GSPC  & 77 & 2022-01-01 & 2022-04-23 & Indices & USA \\
        \bf DJI  & 77 & 2022-01-01 & 2022-04-23 & Indices & USA \\
        \bf IXIC  & 77 & 2022-01-01 & 2022-04-23 &Indices & USA \\ 
        \bf N225  & 77 & 2022-01-01 &2022-04-23  & Indices & JAPAN \\ 
        \bf HSI   & 77 & 2022-01-01 & 2022-04-23 & Indices & HONG KONG \\    
          
        \bf AAPL  & 77 & 2022-01-01 &2022-04-23 &Technology & USA \\
        \bf MSFT  & 77 & 2022-01-01 & 2022-04-23 &Technology & USA \\
        
        \bf AMZN  & 77 & 2022-01-01& 2022-04-23 &E-Commerce & USA \\ 
        \bf BABA  & 77 &2022-01-01& 2022-04-23 &E-Commerce & CHINA \\
        
        \bf TSLA  & 77 & 2022-01-01& 2022-04-23  & Automakers & USA \\ 
        
        \hline    
        \end{tabular}%
        }
      
    \vspace{-0.2in}
    \label{tab:llm_confg2}

\end{table}

The daily financial time series data from January 01, 2022, through April 23, 2022, were collected from Yahoo Finance\footnote{https://finance.yahoo.com/}. The dataset, described in Table \ref{tab:llm_confg2}, includes a diverse selection of stocks and indices. Each dataset is a stock or index with a number of data points, date range, sector and country of origin. Stocks were chosen based on market capitalization and sector representation. As this table \ref{tab:llm_confg2} shows, the selected datasets represent a variety of sectors (indices, technology, e-commerce and automakers), and countries (USA, Japan, Hong Kong, China). This table \ref{tab:llm_confg2} provides a basis to test the LLM-generated models in datasets in terms of variety and geographical diversity.

In aaddition, major indices like the S$\&$P 500 (GSPC) and Dow Jones Industrial Average (DJI), Nasdaq Composite (IXIC), Nikkei 225 (N225) and Hang Seng Index (HSI) were included. Giant hightech companies such as Apple (AAPL) and Microsoft (MSFT) were also included in the dataset. furthermore, E-commerce giants such as Amazon (AMZN) and Alibaba (BABA) were incorporated in the dataset to make the dataset more diverse. Lastly, Automakers like Tesla (TSLA) were also added with the goal of investigating the performance of each model generated by LLMs for each industry sector.

\subsection{LLMs Configuration}
The granularity and diversity of responses generated by LLMs can be controlled via several configuration parameters: 
\begin{enumerate}
\item Temperature control randomization of the responses generated by a large language model where temperature score close to 1 indicates increases in the randomness; 
\item The $top\_p$ parameter filters the predicted next words by computing a cumulative probability distribution. A $top\_p$ value of $0.4$ indicates considering the 40 most likely words or phrases;
\item The $max\_token\_size$ value of $2,048$ limits the large language model analysis to the most recent $2,048$ tokens when generating responses.
\end{enumerate}

\begin{table}[h!]
    \centering
    \caption{{Comparison of Parameters and Settings for Different LLMs.}}
    \scalebox{0.75}{
    \begin{tabular}{|l|c|c|c|c|}
        \hline
        \textbf{Parameter}&\textbf{GPT-3.5-Turbo} &\textbf{Falcon} &\textbf{LLama-2} &\textbf{PaLM} \\        
        \hline

        model & gpt$-$3.5$-$turbo &7B  &7B  & text$-$bison$-$001 \\ \hline 
        temperature & 0.7 &0.7  &0.01  &0.25  \\ \hline 
        max token$\_$size & 1024 &1024  &1024  &1024 \\ \hline
        top$\_$p& 0.7 &0.3  &0.9  &0.095  \\ \hline

    \end{tabular}
   }
    \label{tab:llm_confg}
\end{table}

Table \ref{tab:llm_confg} shows the configuration we used for each model in this experiment. The model represents the version of the base model used with 7B (7 billion) parameters. The values reported in Table \ref{tab:llm_confg} are obtained through fine tuning stages performed through several experiments. For the purposes of our experiments, various temperature settings were used to observe their effects on the model’s performance. For instance, when a higher temperature was used, the model worked well but the formulated hypotheses contained a lot of variability and were not very coherent. Conversely, the low-temperature output settings appeared to be more standardized and dependable, but less varied than the previous higher-temperature outputs. They are taken into account in the interpretation of the results when the diverse experiments were performed to choose the value in Table \ref{tab:llm_confg}.

\subsection{Prompt Engineering with Sensitivity Levels}
The purpose of this research paper is to investigate 1) whether the deep learning-based models generated by LLMs for forecasting time series data are good enough (i.e., comparable with fine-tuned models manually produced by expert data analysts; and 2) whether it is feasible to enhance the goodness of models by controlling the criteria for effective prompt engineering including 1) clarity and specificity, 2) objectives and intents, 3) contextual information, and 4) format and stylealong with sensitive levels of low, medium, and high.

To perform such controlled experiment, the study consider the following criteria in crafting prompts along with additional controlled imposed by sensitivity levels considered.

\textbf{I)  Clarity and Specificity (CS)} of prompts provided to LLMs.
\begin{enumerate}
    \item {\it Low.} Use of vague terms, lack of specific details, and general language that does not clearly convey the desired tasks.
    \item {\it Medium.} Check for clear tasks with some specific details, but might have some rooms for further clarification.
    \item  {\it High.} Identify prompts with well-defined, precise tasks that leave little ambiguity.
\end{enumerate}

\textbf{II) Objectives and Intents (OI) } of prompts provided to LLMs.
\begin{enumerate}
    \item {\it Low.} The prompt's exact objectives and intents are unclear due to the use of ambiguous language or a lack of clear purpose.
    \item {\it Medium.} The prompt states a general objective but does not provide a clear context for the desirable task.
    \item {\it High.} The prompts are crafted with clear statements of objective and intent, providing a well-defined context.
\end{enumerate}

\textbf{III) Contextual Information (CI)} of prompts provided to LLMs.
\begin{enumerate}
    \item {\it Low.} The prompts are provided with minimal contextual information about the data, the problem domain, or the use case, making it hard to understand the request.
    \item {\it Medium.} Identify prompts that offer some context but might lack crucial information about the scenario.
    \item {\it High.} The prompts provide detailed context, including metadata , problem domain, and use case information.
\end{enumerate}

\textbf{IV) Format and Style (FS)} of prompts provided to LLMs.
\begin{enumerate}
    \item {\it Low.} The prompts are written with unclear or inconsistent format explaining the desired code or model.
    \item {\it Medium.} Check for prompts that are generally structured but might have some issues with clarity, style, or terminology.
    \item {\it High.} The prompts are crafted with well-structured language, proper format for desirable code, and appropriate use of terminology.
\end{enumerate}

\subsection{Performance Metric}
Throughout the experiment, the performance metric utilized is the Root-Mean-Square Error (RMSE). This metric serves to assess the output generated by all LLMs in response to the provided input prompts. The RMSE essentially computes the square root of the mean of the squared differences between the model's predictions and the true values. A lower RMSE value indicates that the underlying models had achieved better performances. More specifically, RMSE values indicate that the model's predictions are closer to the actual ground target values on average.

\vspace{-0.2cm}
\section{Experimental Procedure}
\label{sec:expermintal_study}

To assess the performance of LLMs in generating good deep learning-based models for analyzing time series data, we crafted a set of prompts according to the criteria and sensitivity levels discussed earlier. Table \ref{ref:prompts_1} lists each prompt that is crafted according to criteria and sensitivity levels. The prompts are designed and provided to mainstream LLMs as inputs for various levels of sensitivity. We will then provide these prompts to each LLM, take the response of each LLM, compile and execute the code generated by LLMs and capture RMSE for comparison purposes. 

\subsection{Sensitivity Analysis for Designing Prompts}
We crafted eleven prompts ranging from easy to complex sensitivity. The designing of these eleven prompts was based on pair-wise sensitivity analysis where a factor is changing, and remaining factors are kept constant. Pair-wise analysis, also known as pairwise comparison, is a method for comparing and evaluating many items or criteria by comparing each item to every other item in a methodical and systematic manner. The phrase ``pair-wise analysis'' refers to the process of analyzing and comparing the distinct criterion levels (i.e., Low, Medium, and High) against each other for each individual element in the context of the information and determining their impact on the results.
 
The pair-wise analysis helps in evaluating the quality, significance, or applicability of multiple characteristics by directly comparing them to one another, allowing for a more systematic and thorough review process.

\begin{table*}
    \footnotesize
    \centering
    \caption{{Prompts and Colored Categorical Sensitivity Levels (Green: high; Orange: Medium; Red: Low)}}

    \label{ref:prompts_1}
    \scalebox{0.8}{
    \begin{tabularx}{1.25\linewidth}{|c|X|c|c|c|c|}
      \hline
      \multicolumn{1}{|c|}{\bf Prompt} & \multicolumn{1}{c|}{\bf Description} & \textbf{Clarity and } & \textbf{Objective}& \textbf{Contextual} & \textbf{Format}\\
     & & \textbf{Specificity} & \textbf{and Intent}& \textbf{ Information} & \textbf{and Style}\\
    \textbf{} & \textbf{} & \textbf{[CS]} & \textbf{[OI]}& \textbf{[CI]} & \textbf{[FS]}\\
      \hline
      1 &\textcolor{highcolor}{Can you assist me in creating a comprehensive Python script to build an LSTM architecture using the time series dataset enclosed within double backticks \texttt{``\{data\}``}?. My objective is to execute steps such as preprocessing, splitting, building, compiling, training, and evaluating models using  RMSE.}& High & High & High & High\\ 
      \hline
      2 & \textcolor{mediumcolor}{Could you assist me in generating a Python script to build an LSTM model using the provided time series dataset enclosed within double backticks \texttt{``\{data\}``}? My goal is to perform preprocessing, splitting the given data, creating the model, compiling it, training the model, and assessing its performance using RMSE.}& Medium & Medium & Medium & Medium\\
      \hline
      3 & \textcolor{lowcolor}{I need a Python script for LSTM. The dataset is in  \texttt{``\{data\}``}.I want to process, split, build, compile, train model, and evaluate model.}& Low & Low & Low & Low\\
    \hline
    
      4 & \textcolor{highcolor}{Could you \textcolor{mediumcolor}{give me a code for setting up a LSTM?} I have a time series dataset enclosed within double backticks \texttt{``\{data\}``}. My goal is to process the data, split it, build the model, compile it, train, and evaluate using RMSE.}& Medium & High & High & High\\
    \hline
      5 & \textcolor{highcolor}{Could you  \textcolor{lowcolor}{ help me out with crafting some kind of Python code} to establish an LSTM architecture using the enclosed within double backticks \texttt{``\{data\}``}? \textcolor{lowcolor}{ To execute thorough preprocessing, split, build, compilation, training, and evaluation.}}& Low & High & High & High\\
    \hline
    
      
      6 & \textcolor{highcolor}{Can you help me with creating \textcolor{mediumcolor}{a Python script for an LSTM architecture} using the time series dataset enclosed within double backticks \texttt{``\{data\}``}? \textcolor{mediumcolor}{If possible I  would like} to perform preprocessing, data splitting, model construction, compilation, training, and evaluation using RMSE using the code.} & High & Medium & High & High
     \\
     \hline
      7 & \textcolor{highcolor}{Could you maybe assist me \textcolor{lowcolor}{with making a Python script} \textcolor{highcolor}{to create an LSTM architecture using the time series dataset enclosed within double backticks \texttt{``\{data\}``}?}  \textcolor{lowcolor}{ To perform} preprocessing, splitting, building, compiling, training, and testing using RMSE.}& High & Low & High & High
     \\
    \hline
      8 & \textcolor{highcolor}{Can you help me \textcolor{mediumcolor}{to establish an LSTM architecture in Python using the enclosed within double backticks \texttt{``\{data\}``} to forecast stock prices?} My aim is to perform thorough \textcolor{mediumcolor}{preprocessing, divide the data, construct the model,  and evaluate its performance using RMSE.}}& High & High & Medium & High
     \\
     \hline
      9 &\textcolor{highcolor}{ \textcolor{lowcolor}{Could you
      help me} in making a comprehensive Python script \textcolor{lowcolor}{to build an LSTM architecture using the  dataset \texttt{``\{data\}``}.} My aim is to execute carefully \textcolor{lowcolor}{preprocessing,  construct the architecture,  and assess performance using RMSE.}}& High & High & Low & High
    \\
    \hline
      10 & \textcolor{highcolor}{Would you be able to help me in generating a  \textcolor{mediumcolor}{Python to set an LSTM architecture} using the time series dataset enclosed within double backticks \texttt{``\{data\}``}? My \textcolor{mediumcolor}{steps include }  preprocessing, dividing, constructing, compiling, training, and evaluating using RMSE.}& High & High & High & Medium
     \\
     \hline
     11& \textcolor{highcolor}{Could you please} \textcolor{lowcolor}{help me with generating a script} \textcolor{highcolor}{to build an LSTM architecture using the time series dataset enclosed within double backticks \texttt{``\{data\}``}?} \textcolor{lowcolor}{Perform preprocessing, division of data, construction of the model, compilation, training, and evaluation the model.}& High & High & High & Low
    \\
      \hline
    \end{tabularx}
    }
\end{table*}

To help trace the sensitivity levels, a coloring scheme is employed where the green, orange, and red colors in Table \ref{ref:prompts_1} represent sensitivity level of high, medium, and low, respectively. 

\subsection{Manual Creation and Optimization a Model}
The experiments execute on Apple M1 MAX, Memory of 64 with GPU. The dataset split into 80\% for training and 20\% testing for testing. In the preprocessing, the data is scaled using \textit{MinMaxscaler}, which transforms the feature range from 0 to 1 where the data linearly scales down. After the scaling, the data is prepared into sequences of length 5 to predict the next day's (1) data and feed into the model for training. The manual creation of the model consists of the LSTM architecture with tensorflow in the backend. The model contains One LSTM layer with 50 Units with 'relu' activation function. The model trains with 100 epochs with a batch size of 1. The model employs 'adam' as an optimizer and 'mse' as loss function. The hyperparameters were chosen with various observations during the experiments. The preprocessing steps and building model are only relevant for the manual creation, as LLMs are provided the raw data for code generation.

\section{Results}
\label{sec:results}
Table \ref{tab:result1} reports the results the performance of the deep learning-based models generated by LLMs for time series data analysis. Each model is evaluated using RMSE values for each stock data. The PaLM model achieved the lowest RMSE value of 0.0023 for BABA ticker while the Falcon achieved the lowest RMSE value of 0.0041 for the GSPC ticker. The LLama2 did not achieve the lowest RMSE across all tickers,  whereas the GPT 3.5 has the lowest RMSE for eight tickers. However, the manually developed and optimized model achieved the lowest RMSE compared to LLM generated model across all tickers.

\begin{table*}
\footnotesize
    \caption{{RMSE values for Models Generated Using LLMs and Controlled Prompts, with LLM configurations Detailed in Table \ref{tab:llm_confg} and Prompts in Table \ref{ref:prompts_1}}.\\}

    \label{tab:result1}
    \vspace{-0.3cm}
    \begin{subtable}{0.5\linewidth}
      \centering
        \scalebox{0.75}{
       \begin{tabular}{|c|c|c|c|c|c|}
          \hline
          \multirow{2}{*}{\textbf{Ticker}} & \multirow{2}{*}{\textbf{Prompts}} & \multicolumn{4}{c|}{\textbf{RMSE}} \\
         \cline{3-6}
            &  & \textbf{PaLM} & \textbf{falcon} & \textbf{LLama 2} & \textbf{GPT 3.5} \\
          \hline
      
                 \multirow{10}{*}{\textbf{GSPC}} 
                    & 1 & 0.0323 & 0.0331 & 0.0389 & NA  \\
                    \cline{2-6}
                    & 2 & 0.0368 & 0.4893 & 0.0394 & 0.0479  \\
                    \cline{2-6}
                    & 3 & 0.0388 &0.0359 & 0.0413 & NA \\
                    \cline{2-6}
                    & 4 & 0.0318 & 0.1992 &0.0367 & NA  \\
                    \cline{2-6}
                    & 5 & \cellcolor{lightgray} \bf{0.0216} & NA & 0.0410 & 0.0633  \\
                    \cline{2-6}
                    & 6 & 0.0314 & 0.0356 & 0.0376 & NA \\
                    \cline{2-6}
                    & 7 & 0.0348 & \cellcolor{black} \textcolor{white} {\bf 0.0041} & 0.0390 & NA \\
                    \cline{2-6}
                    & 8 & 0.0331 & 0.4649 & \cellcolor{lightgray}  {\bf 0.0330} & 0.1043 \\
                    \cline{2-6}
                    & 9 & 0.0320 & NA & 0.0456 &0.0411 \\
                    \cline{2-6}
                    & 10 & 0.0335 &0.0355 & 0.0381 & \cellcolor{lightgray}  {\bf 0.0376}  \\
                    \cline{2-6}
                    & 11 & 0.0353 & 0.0905 & 0.0429 & NA  \\
                    \cline{2-6}
                    & Avg. & \bf \avercalc[11]{0.0323, 0.0368, 0.0388, 0.0318, 0.0216, 0.0314, 0.0348, 0.0331,0.0320,0.0335,0.0353} 
                    &\bf \avercalc[9]{0.0331, 0.4893,0.0359, 0.1992,0.0356,0.0041,0.4649,0.0355,0.0905}
                    & \bf \avercalc[11]{0.0389, 0.0394,0.0413,0.0367,0.0410,0.0376,0.0390,0.0330,0.0456,0.0381,0.0429}
                    & \bf \avercalc[5]{0.0479,0.0633,0.1043,0.0411,0.0375}
                    \\
                    \cline{2-6}
                    & \cellcolor{lightgray} Manual & \multicolumn{4}{c|}{\cellcolor{lightgray} \bf Manually Developed \& Optimized Model: 0.0058} \\
                    \cline{2-6}
                    \hline
            
                \multirow{10}{*}{\textbf{DJI}} 
                    & 1 & 0.0386 & NA & \cellcolor{lightgray} {\bf 0.0399} & 0.0464  \\
                    \cline{2-6}
                    & 2 & 0.0392 & NA & 0.0556 & 0.2476  \\
                    \cline{2-6}
                    & 3 & 0.0444 & 0.4935 & 0.0462 & 0.3364 \\
                    \cline{2-6}
                    & 4 & 0.0425 & \cellcolor{lightgray} {\bf 0.0409} & 0.0439 & 0.0567  \\
                    \cline{2-6}
                    & 5 & 0.0385 & NA & 0.0472 & 0.0555  \\
                    \cline{2-6}
                    & 6 & 0.0438 & NA & 0.0445 & NA \\
                    \cline{2-6}
                    & 7 & 0.0395 & NA & 0.0425 & NA \\
                    \cline{2-6}
                    & 8 & \cellcolor{lightgray} {\bf 0.0365} & 0.4642 & 0.0413 & 0.2446 \\
                    \cline{2-6}
                    & 9 & 0.0394 & NA & 0.0487 & \cellcolor{black} \textcolor{white} {\bf 0.0097} \\
                    \cline{2-6}
                    & 10 & 0.0417 & NA & 0.0473 & 0.2960  \\
                    \cline{2-6}
                    & 11 & 0.4061 & NA & 0.0544 & NA  \\
                    \cline{2-6}
                    & Avg. & \bf \avercalc[11]{0.0386,0.0392,0.0444,0.0425,0.0385,0.0438,0.0395,0.0365,0.0394,0.0417,0.4061}
                    & \bf \avercalc[3]{0.4935,0.0409,0.4642} 
                    & \bf \avercalc[11]{0.0399, 0.0556, 0.0462,0.0439,0.0472,0.0445, 0.0425,0.0413,0.0487,0.0473,0.0544}
                    & \bf \avercalc[8]{0.0464, 0.2476, 0.3364, 0.0567, 0.0555, 0.2446,0.0097,0.2960} \\
                    
                    \cline{2-6}
                    & \cellcolor{lightgray} Manual & \multicolumn{4}{c|}{\cellcolor{lightgray} \bf Manually Developed \& Optimized Model: 0.0053} \\
                    \cline{2-6}
                    \hline
                
                \multirow{10}{*}{\textbf{IXIC}} 
                   & 1 & \cellcolor{lightgray} {\bf 0.0259} & NA & 0.0313 & 0.1000  \\
                    \cline{2-6}
                    & 2 & 0.0285 & NA & 0.0324 & NA  \\
                    \cline{2-6}
                    & 3 & 0.0284 & NA & 0.0371 & NA \\
                    \cline{2-6}
                    & 4 & 0.0303 & \cellcolor{lightgray} {\bf 0.0319} & 0.0327 & NA  \\
                    \cline{2-6}
                    & 5 & 0.0300 & 0.1574 & 0.0311 & 0.0325  \\
                    \cline{2-6}
                    & 6 & 0.0289 & 0.3752 & 0.0316 & NA \\
                    \cline{2-6}
                    & 7 & 0.0299 & NA & 0.0348 & NA \\
                    \cline{2-6}
                    & 8 & 0.0343 & 0.4024 & \cellcolor{lightgray} {\bf 0.0285} & 0.0310  \\
                    \cline{2-6}
                    & 9 & 0.0299 & 0.2674 & 0.0323 & NA \\
                    \cline{2-6}
                    & 10 & 0.0294 & NA & 0.0321 & \cellcolor{black} \textcolor{white} {\bf 0.0041}  \\
                    \cline{2-6}
                    & 11 & 0.0286 & NA & 0.0316 & NA  \\
                    \cline{2-6}
                    & Avg. & \bf \avercalc[11]{0.0259,0.0285,0.0284,0.0303,0.0300,0.0289,0.0299,0.0343,0.0299,0.0294,0.0286} 
                    & \bf \avercalc[5]{0.0319,0.1574,0.3752,0.4024,0.2674} 
                    & \bf \avercalc[11]{0.0313,0.324,0.0371,0.0327,0.0311,0.0316,0.0348,0.0285,0.0323,0.0321,0.0316}
                    & \bf \avercalc[4]{0.1000,0.0325,0.0310,0.0041} \\
                    \cline{2-6}
                    & \cellcolor{lightgray} Manual & \multicolumn{4}{c|}{\cellcolor{lightgray} \bf Manually Developed \& Optimized Model: 0.0070} \\
                    \cline{2-6}
                    \hline
                
                \multirow{10}{*}{\textbf{N225}} 
                    & 1 & NA & 0.0801 & 0.0317 & NA  \\ 
                    \cline{2-6}
                    & 2 & 0.0291 & NA & 0.0319 & NA  \\
                    \cline{2-6}
                    & 3 & NA & NA & 0.0358 & 0.0286 \\
                    \cline{2-6}
                    & 4 & 0.0307 &0.3902 & 0.0325 & NA  \\ 
                    \cline{2-6}
                    & 5 & 0.0209 & 0.4513 & 0.0424 & 0.1568  \\
                    \cline{2-6}
                    & 6 & 0.0353 & NA & 0.0412 & 0.0355 \\
                    \cline{2-6}
                    & 7 & \cellcolor{lightgray} {\bf 0.0182} & NA &0.0465 & NA \\
                    \cline{2-6}
                    & 8 & NA & NA & \cellcolor{lightgray} {\bf 0.0305} & 0.0281  \\ 
                    \cline{2-6}
                    & 9 & 0.0351 & NA & 0.0432 & \cellcolor{black} \textcolor{white} {\bf 0.0039} \\
                    \cline{2-6}
                    & 10 & 0.0323 & \cellcolor{lightgray} {\bf 0.0199} & 0.0348 & 0.0049  \\
                    \cline{2-6}
                    & 11 &0.0415 & NA & 0.0322 & NA  \\
                    \cline{2-6}
                    & Avg. & \bf \avercalc[8]{0.0291,0.0307,0.0209,0.0353,0.0182,0.0351,0.0323,0.0415}
                    & \bf \avercalc[4]{0.0801,0.3902,0.4513,0.0199} 
                    & \bf \avercalc[11]{0.0317, 0.0319, 0.0358, 0.0325, 0.0424, 0.0412, 0.0465, 0.0305, 0.0432, 0.0348, 0.0322}
                    & \bf \avercalc[6]{0.0286, 0.1568, 0.0355, 0.0281, 0.0039, 0.0049} \\
                    \cline{2-6}
                    & \cellcolor{lightgray} Manual & \multicolumn{4}{c|}{\cellcolor{lightgray} \bf Manually Developed \& Optimized Model: 0.0034} \\
                    \cline{2-6}
                    \hline
    
                \multirow{10}{*}{\textbf{HSI}} 
                    & 1 & NA & \cellcolor{lightgray} {\bf 0.3867} & 0.0183 & 0.0067  \\ 
                    \cline{2-6}
                    & 2 & NA & 0.4212 & 0.0193 &0.0073 \\
                    \cline{2-6}
                    & 3 & NA & NA &0.0307 & NA  \\ 
                    \cline{2-6}
                    & 4 & NA & NA & 0.0196 & 0.0056 \\
                    \cline{2-6}
                    & 5& NA & NA & 0.0271 & 0.0212 \\
                    \cline{2-6}
                    & 6 & NA & NA & 0.0193 & \cellcolor{black} \textcolor{white} {\bf 0.0053} \\
                    \cline{2-6}
                    & 7 & NA &0.5887 & \cellcolor{lightgray} {\bf 0.0171} & 0.0274 \\
                    \cline{2-6}
                    & 8 & 0.0455 & NA & 0.0259 & 0.0278  \\
                    \cline{2-6}
                    & 9 & 0.0280 & NA & 0.0488 & 0.0413 \\
                    \cline{2-6}
                    & 10 & NA & NA & 0.0179 & 0.0178 \\
                    \cline{2-6}
                    & 11 & \cellcolor{lightgray} {\bf 0.0196} & NA & 0.0283 & NA  \\
                    \cline{2-6}
                    & Avg. & \bf \avercalc[3]{0.0455, 0.280, 0.0196} 
                    & \bf \avercalc[3]{0.3867,0.4212,0.5887} 
                    & \bf \avercalc[11]{0.0183, 0.0193, 0.0307, 0.0196, 0.0271, 0.0193, 0.0171, 0.0259, 0.0488, 0.0179, 0.0283} 
                    & \bf \avercalc[9]{0.0067, 0.0073, 0.0056, 0.0212, 0.0053, 0.0274, 0.0278, 0.0413, 0.0178} \\
                    \cline{2-6}
                    & \cellcolor{lightgray} Manual & \multicolumn{4}{c|}{\cellcolor{lightgray} \bf Manually Developed \& Optimized Model: 0.0354} \\
                    \cline{2-6}
                    \hline
              \end{tabular}
          }
    \end{subtable}%
    \begin{subtable}{0.5\linewidth}
      \centering
    \scalebox{0.75}{
    \begin{tabular}{|c|c|c|c|c|c|}
      \hline
      \multirow{2}{*}{\textbf{Ticker}} & \multirow{2}{*}{\textbf{Prompts}} & \multicolumn{4}{c|}{\textbf{RMSE}} \\
     \cline{3-6}
        &  & \textbf{PaLM} & \textbf{falcon} & \textbf{LLama 2} & \textbf{GPT 3.5} \\
      \hline
  
\multirow{10}{*}{\textbf{AAPL}} 
        & 1 & 0.0360 & NA & 0.0390  & 0.0384  \\
        \cline{2-6}
        & 2 & 0.0365 & \cellcolor{lightgray} {\bf 0.5292} & 0.0403  & 0.0417  \\
        \cline{2-6}
        & 3 & 0.1039 & NA & 0.0426 & 0.0483 \\
        \cline{2-6}
        & 4 & 0.0369 & NA& 0.0392  & 0.0404  \\
        \cline{2-6}
        & 5 & 0.1096 & 0.6225& 0.0432  & 0.1135  \\
        \cline{2-6}
        & 6 & 0.0373 & NA& 0.0382  & 0.0396 \\
        \cline{2-6}
        & 7 & 0.0366 & 0.6267& 0.0412  &  0.0962 \\
        \cline{2-6}
        & 8 & \cellcolor{lightgray} {\bf 0.0351} & NA& \cellcolor{lightgray} {\bf 0.0364}  & 0.1745  \\
        \cline{2-6}
        & 9 & 0.0366 & NA& 0.0408  & 0.1221 \\
        \cline{2-6}
        & 10 & 0.0379 & 0.7085& 0.0397  & \cellcolor{black} \textcolor{white} {\bf 0.0062}  \\
        \cline{2-6}
        & 11 & 0.0381 & NA& 0.0409 & 0.0641  \\
                    \cline{2-6}
         & Avg. 
         & \bf \avercalc[11]{0.0360, 0.0365, 0.1039, 0.0369, 0.1096, 0.0373, 0.0366, 0.0351, 0.0366, 0.0379, 0.0381} 
         & \bf \avercalc[4]{0.5292, 0.6225, 0.6267, 0.7085} 
         & \bf \avercalc[11]{0.0390, 0.0403, 0.0426, 0.0392, 0.0432, 0.0382, 0.0412, 0.0364, 0.0408, 0.0397, 0.0409} 
         & \bf \avercalc[11]{0.0384, 0.0417, 0.0483, 0.0404, 0.1135, 0.0396, 0.0962, 0.1745, 0.1221, 0.0062, 0.0641} \\
        \cline{2-6}
        & \cellcolor{lightgray} Manual & \multicolumn{4}{c|}{\cellcolor{lightgray} \bf Manually Developed \& Optimized Model: 0.0094} \\
        \cline{2-6}
        \hline
\multirow{10}{*}{\textbf{MSFT}} 
        & 1 & 0.0407 & 0.0618 & 0.0415 & 0.0918  \\
        \cline{2-6}
        & 2 & 0.0414 & NA & 0.0686 & 0.1749  \\
        \cline{2-6}
        & 3 & 0.1154 & NA & 0.0423 & 0.2328 \\
        \cline{2-6}
        & 4 & 0.0400 & \cellcolor{lightgray} {\bf 0.0045} & 0.0505 & \cellcolor{black} \textcolor{white} {\bf 0.0041}  \\
        \cline{2-6}
        & 5 & 0.2358 & NA & 0.0417 & 0.2681  \\
        \cline{2-6}
        & 6 & 0.0389 & 0.3535 & 0.0476 & 0.1263 \\
        \cline{2-6}
        & 7 & 0.0391 & NA & 0.0485 & 0.1219 \\
        \cline{2-6}
        & 8 & 0.0369 & NA & \cellcolor{lightgray} {\bf 0.0375} & 0.1811  \\
        \cline{2-6}
        & 9 & 0.0361 & 0.0405 & 0.0390 & 0.1217 \\
        \cline{2-6}
        & 10 & 0.0403 & NA & 0.0425 & 0.1368  \\
        \cline{2-6}
        & 11 & \cellcolor{lightgray} {\bf 0.0343} & 0.3160 & 0.0426 & 0.1170  \\
                    \cline{2-6}
        & Avg. 
        & \bf \avercalc[11]{0.0407, 0.0414, 0.1154, 0.0400, 0.2358, 0.0389, 0.0391, 0.0369, 0.0361, 0.0403, 0.0343} 
        & \bf \avercalc[5]{0.0618, 0.0045, 0.3535, 0.0405, 0.3160}
        & \bf \avercalc[11]{0.0415, 0.0686, 0.0423, 0.0505, 0.0417, 0.0476, 0.0485, 0.0375, 0.0390, 0.0425, 0.0426}
        &\bf \avercalc[11] {0.0918, 0.1749, 0.2328, 0.0041, 0.2681, 0.1263, 0.1219, 0.1811, 0.1217, 0.1368, 0.1170} \\
        \cline{2-6}
        & \cellcolor{lightgray} Manual & \multicolumn{4}{c|}{\cellcolor{lightgray} \bf Manually Developed \& Optimized Model: 0.0065} \\
        \cline{2-6}
        \hline
\multirow{10}{*}{\textbf{AMZN}} 
        & 1 & 0.0421 & 0.6003 & 0.0463 & 0.0462  \\
        \cline{2-6}
        & 2 & 0.0430 & \cellcolor{lightgray} {\bf 0.5393} & 0.0480 & 0.1571  \\
        \cline{2-6}
        & 3 & 0.0797 & NA & 0.0451 & 0.0481 \\
        \cline{2-6}
        & 4 & 0.0432 & NA & 0.0455 & 0.0445  \\
        \cline{2-6}
        & 5 & 0.1146 & 0.5977 & 0.0430 & 0.0814  \\
        \cline{2-6}
        & 6 & 0.0423 & NA & 0.0447 & 0.0727 \\
        \cline{2-6}
        & 7 & 0.1632 & NA & 0.0478 & 0.0813 \\
        \cline{2-6}
        & 8 & \cellcolor{lightgray} {\bf 0.0400} & NA & \cellcolor{lightgray} {\bf 0.0375} & 0.1673  \\
        \cline{2-6}
        & 9 & 0.0432 & NA & 0.0472 & \cellcolor{black} \textcolor{white} {\bf 0.0277} \\
        \cline{2-6}
        & 10 & 0.0424 & NA & 0.0435 & 0.0704  \\
        \cline{2-6}
        & 11 & 0.0417 & NA & 0.0439 & 0.0797  \\
                    \cline{2-6}
        & Avg. 
        & \bf \avercalc[11]{0.0421, 0.0430, 0.0797, 0.0432, 0.1146, 0.0423, 0.1632, 0.0400, 0.0432, 0.0424, 0.0417}
        & \bf \avercalc[3]{0.6003, 0.5393,0.5977} 
        & \bf \avercalc[11]{0.0463, 0.0480, 0.0451, 0.0455, 0.0430, 0.0447, 0.0478, 0.0375, 0.0472, 0.0435, 0.0439} 
        & \bf \avercalc[11]{0.0462, 0.1571, 0.0481, 0.0445, 0.0814, 0.0727, 0.0813, 0.1673, 0.0277, 0.0704, 0.0797} \\
        \cline{2-6}
        & \cellcolor{lightgray} Manual & \multicolumn{4}{c|}{\cellcolor{lightgray} \bf Manually Developed \& Optimized Model: 0.0062} \\

        \cline{2-6}
        \hline
\multirow{10}{*}{\textbf{BABA}} 
        & 1 & 0.0238 & 0.3414 & 0.0346 & 0.0864  \\
        \cline{2-6}
        & 2 & 0.0241 & NA & 0.0392 & 0.0871  \\
        \cline{2-6}
        & 3 & 0.1494 & NA & 0.0389 & 0.2412  \\
        \cline{2-6}
        & 4 & 0.0257 & NA & 0.0366 & 0.0370  \\
        \cline{2-6}
        & 5 & 0.2449 & 0.0486 & 0.0288 & 0.0659  \\
        \cline{2-6}
        & 6 & 0.0243 & 0.0673 & 0.0300 & 0.1559 \\
        \cline{2-6}
        & 7 & 0.0242 & 0.3419 & 0.0359 & 0.0623 \\
        \cline{2-6}
        & 8 & \cellcolor{black} \textcolor{white} {\bf 0.0223} & NA & \cellcolor{lightgray} {\bf 0.0246} & 0.2253  \\
        \cline{2-6}
        & 9 & 0.0246 & NA & 0.0256 & NA \\
        \cline{2-6}
        & 10 & 0.0233 & \cellcolor{lightgray} {\bf 0.0237} & 0.0437 & \cellcolor{lightgray} {\bf 0.0286}  \\
        \cline{2-6}
        & 11 & 0.0228 & NA & 0.0631 & 0.0674  \\
                    \cline{2-6}
        & Avg. 
        & \bf \avercalc[11]{0.0238, 0.0241, 0.1494, 0.0257, 0.2449, 0.0243, 0.0242, 0.0223, 0.0246, 0.0233, 0.0228} 
        & \bf \avercalc[5]{0.3414,  0.0486, 0.0673, 0.3419, 0.0237}
        & \bf \avercalc[11]{0.0346, 0.0392, 0.0389, 0.0366, 0.0288, 0.0300, 0.0359, 0.0246, 0.0256, 0.0437, 0.0631} 
        & \bf \avercalc[11]{0.0864, 0.0871, 0.2412, 0.0370, 0.0659, 0.1559, 0.0623, 0.2253, 0.0286, 0.0674} \\
        \cline{2-6}
        & \cellcolor{lightgray} Manual & \multicolumn{4}{c|}{\cellcolor{lightgray} \bf Manually Developed \& Optimized Model: 0.0268} \\

        \cline{2-6}
        \hline
\multirow{10}{*}{\textbf{TSLA}} 
        & 1 & 0.0345 & NA & 0.0424 & 0.0466  \\
        \cline{2-6}
        & 2 & 0.0347 & 0.1398 & 0.0403 & 0.1009  \\
        \cline{2-6}
        & 3 & 0.5289 & NA & 0.0564 & 0.0027 \\
        \cline{2-6}
        & 4 & 0.0376 & 0.0359 & 0.0414 & 0.0044  \\
        \cline{2-6}
        & 5 & 0.0354 & NA & 0.0519 & \cellcolor{black} \textcolor{white} {\bf 0.0023}  \\
        \cline{2-6}
        & 6 & 0.0361 & \cellcolor{lightgray} {\bf 0.0048} & 0.0373 & 0.0540 \\
        \cline{2-6}
        & 7 & 0.0388 & NA & 0.0390 & 0.0031 \\
        \cline{2-6}
        & 8 & 0.0360 & 0.0556 & \cellcolor{lightgray} {\bf 0.0343} & 0.0820  \\
        \cline{2-6}
        & 9 & \cellcolor{lightgray} {\bf 0.0334} & 0.0356 & 0.0376 & 0.1634 \\
        \cline{2-6}
        & 10 & 0.0363 & 0.0655 & 0.0387 & 0.0785  \\
        \cline{2-6}
        & 11 & 0.0384 & NA & 0.0483 & 0.0897  \\
                    \cline{2-6}
        & Avg. 
        & \bf \avercalc[11]{0.0345, 0.0347, 0.5289, 0.0376, 0.0354, 0.0361, 0.0388, 0.0360, 0.0334, 0.0363, 0.0384} 
        & \bf \avercalc[6]{ 0.1398,  0.0359, 0.0048,  0.0556, 0.0356, 0.0655} 
        & \bf \avercalc[11]{0.0424, 0.0403, 0.0564, 0.0414, 0.0519, 0.0373, 0.0390, 0.0343, 0.0376, 0.0387, 0.0483} 
        & \bf \avercalc[11]{0.0466, 0.1009, 0.0027, 0.0044, 0.0023, 0.0540, 0.0031, 0.0820, 0.1634, 0.0785, 0.0897} \\
        \cline{2-6}
        & \cellcolor{lightgray} Manual & \multicolumn{4}{c|}{\cellcolor{lightgray} \bf Manually Developed \& Optimized Model: 0.0100} \\
        \cline{2-6}
        \hline 
    \end{tabular}
}
    \end{subtable} 
\end{table*}

In Table \ref{tab:result1}, the best RMSE values obtained for each language model is highlighted with gray color. Moreover, the best RMSE values among all models for each stock data and for all 11 different prompts are highlighted in dark color. The cells with NA indicate that the models generated by the underlying LLM were meaningless and the underlying LLMs produced some other types of models such as regression models instead of deep learning-based models for forecasting time series data. In other words, the  NA values represent the output of the LLMs with no code related to the LSTM model or code not related to predicting time series. This might be due to hallucination problem known in language models where the underlying LLM confuses leveraging its trained data to properly responding to queries and prompts. A noticeable case is the falcon case where the number of NA (the irrelevant response to prompts) is outnumbering the expected responses. This may indicate that the falcon large language model is suffering from the hallucination problem more than the other LLMs.

{\it I) The Performance of Generated Models Across LLMs.} As Table \ref{tab:result1} indicates, on average (i.e., the last rows of each stock data) there is no clear winner among the language models for the eleven prompts studied. The deep learning-based models generated by each LLM are rather comparably competitive. However, as Table \ref{tab:result1} shows, the models generated by GPT 3.5 on prompt 9 and 10 outperform the other generated models (i.e., the dar cells in the table) except for the GPSC and BABA stock data. 

{\it II) The Performance of Generated Models Across Prompts.} We observe that for the case of GPT 3.5, the best models with minimal RMSE values are produced by prompts 8, 9, and 10 where three criteria as 1) clarify and specificity, 2) objective and intent, and 3) Format and Style are set high. 

For the case of LLama 2, we observe that the language model generates the best model using prompt 8 where where three criteria as 1) clarify and specificity, 2) objective and intent, and 3) Format and Style are set high (in most cases).  We also observe a similar pattern for models generated by PaLM through prompts 7, 8, and 9. For the models generated by falcon, there is no clear pattern whether any prompt standout in the comparison where the results are mixed. 

While the results and performance of models and prompts are dispersed, we observe a clear pattern where prompts 8 and 9 seem to produce the best results in generating more accurate models for forecasting time series models where  three criteria as 1) clarify and specificity, 2) objective and intent, and 3) Format and Style are set high.

{\it III) The Performance of Generated Models Across the Time Series Datasets.} As Table \ref{tab:result1} and the black cells indicate, the best results across different dataset is produced by GPT 3.5 mostly by prompts 8, 9, and 10. This may indicate that, at least for GPT 3.5, the more clear and specific (CS), and the more objectively crafted prompt with clear intention (OI), and a clear expression regarding the desired output and format (FS) in the prompts will yield creating better and more accurate models for forecasting time series data. 

{\it IV) The Performance of Generated Models and Manually Developed and Optimized Model.} The most important observation is the accuracy of models created and optimized manually in comparison with the models generated by prompts. 

\begin{figure*}
    \centering
    \includegraphics[width=\textwidth, height=8.75cm]{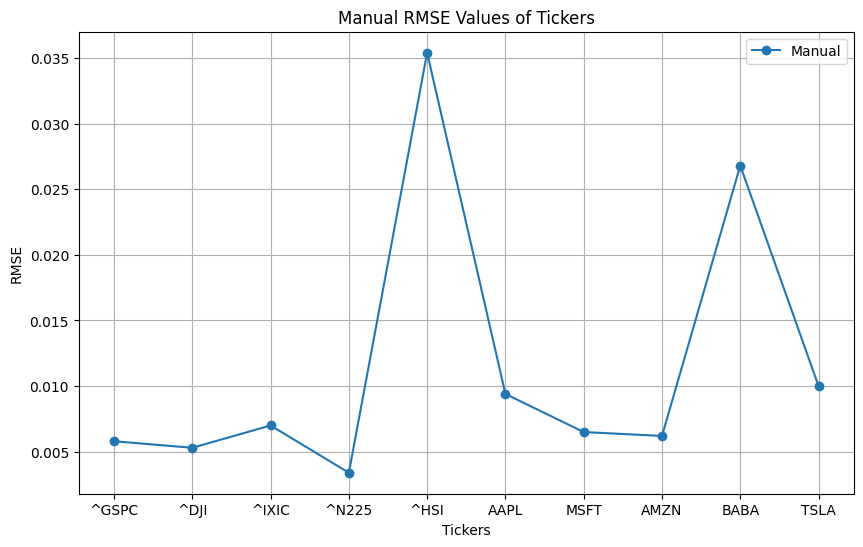}
    \caption{{RMSE Values observed through Manual Implemented LSTM Model.}}
    \label{fig:enter-label}
\end{figure*}

It is important to note that the manually created and optimized model was created based on all data and thus there is only one single manually created and optimized model to compare the results with. More specifically, we did not manually craft and optimize separate deep learning-based models for each dataset. We created a single optimized model for all dataset all together. Figure \ref{fig:enter-label} depicts the RMSE values of the manually crafted and optimized single model obtained for each dataset.  In Figure \ref{fig:enter-label}, the RMSE values of the manually implemented LSTM model on HSI achieved the highest RMSE of value 0.0354 and N225 achieved the lowest RMSE value of 0.0034.

While the manually crafted and optimized model outperforms on three sets of stock data, the models generated by LLMs are also outperforming the manually crafted and optimized models for the seven sets of stock data. More specifically, we observe that the manually created and optimized model outperforms models generated by LLMs for DJI, N225, and AMZN; whereas, the models created through prompts outperform manually created and optimized model for GSPC, IXIC, HSI, AAPL, MSFT, BABA, and TSLA. 

It is important to note that the results are compared based on the best results obtained by the prompts and the variances of RMSE values among different prompts and LLMs are still playing an important indicator in making the final judgment. However, given that prompts 8, 9, and 10 outperform the other prompts in most cases, one case conclude that to generate a comparably good model it is better to set Clarity and Specificity (CS), Objective and Intent (OI), and Format and Style (FS) high and use GPT 3.5 language model to generate the deep learning-based models that can be comparable with manually crafted and optimized model for forecasting time series data.

\subsection{Fixed/Consistent Configurations of LLMs}

The results reported in Table \ref{tab:result1} are based on the configuration and settings of LLMs parameters listed in Table \ref{tab:llm_confg} where each model wes fine-tuned empirically to obtain the best results. To investigate whether various  configuration and parameter settings for LLMs have any effect on the results, we replicate the study with fixed and consistent parameter settings for all LLMs. 

The result in Table \ref{tab:result2} are obtained using the same set of parameters but with consistent and fix values as follows:
1) \texttt{temperature}$ =  0.1$, 2) \texttt{max token\_size}$ =  1,024$ and 3) \texttt{top\_p}$= 0.6$ in all models including GPT 3.5 Turbo, Falcon, Llama-2 and PaLM. This setting primarily means reducing randomness in generating responses to queries or prompts.

\begin{table*}
\footnotesize
    \caption{{RMSE values for Models Generated using  LLMs and Controlled Prompts with fixed LLM configurations 1) \texttt{temperature}$ =  0.1$, 2) \texttt{max token\_size}$ =  1,024$ and 3) \texttt{top\_p}$= 0.6$  and Prompts listed in Table \ref{ref:prompts_1}}.\\}
    \label{tab:result2}
    \vspace{-0.3cm}
    \begin{subtable}{0.5\linewidth}
      \centering
      \scalebox{0.75}{
       \begin{tabular}{|c|c|c|c|c|c|}
          \hline
          \multirow{2}{*}{\textbf{Ticker}} & \multirow{2}{*}{\textbf{Prompts}} & \multicolumn{4}{c|}{\textbf{RMSE}} \\
         \cline{3-6}
            &  & \textbf{PaLM} & \textbf{falcon} & \textbf{LLama 2} & \textbf{GPT 3.5} \\
          \hline
      
                 \multirow{10}{*}{\textbf{GSPC}} 
                 & 1 & 0.0354 & NA & 0.0404 & 0.0409 \\
                \cline{2-6}
                & 2  & 0.0356 & 0.0416 & 0.4802 & \cellcolor{black}\textcolor{white}{\bf0.0036}  \\
                \cline{2-6}
                & 3  & 0.0508 & NA & 0.4663 & 0.0626 \\
                \cline{2-6}
                & 4  & 0.0353 & \cellcolor{lightgray}{\bf 0.0365} & 0.0381 & 0.0756  \\
                \cline{2-6}
                & 5  & 0.2506 & NA & 0.0372 & 0.0964 \\
                \cline{2-6}
                & 6  & 0.0345 & 0.3862 & 0.0377 & 0.0403\\
                \cline{2-6}
                & 7  & 0.0352 & NA & 0.0376 & 0.0641 \\
                \cline{2-6}
                & 8  & 0.0346 & NA & 0.0428 & 0.0403 \\
                \cline{2-6}
                & 9  & 0.0357 & NA & \cellcolor{lightgray}{\bf0.0365} & 0.0617 \\
                \cline{2-6}
                & 10  & \cellcolor{lightgray}{\bf 0.0343} & NA & 0.0490 & 0.1311  \\
                \cline{2-6}
                & 11  & 0.0354 & NA & 0.0427 & 0.0627 \\
                \cline{2-6}
                 & Avg. 
                 & \bf \avercalc[11]{0.0354, 0.0356, 0.0508, 0.0353, 0.2506, 0.0345, 0.0352, 0.0346, 0.0357, 0.0343, 0.0354} 
                 & \bf \avercalc[3]{0.0416, 0.0365,  0.3862} 
                 & \bf \avercalc[11]{0.0404, 0.4802, 0.4663, 0.0381, 0.0372, 0.0377, 0.0376, 0.0428, 0.0365, 0.0490, 0.0427} 
                 & \bf \avercalc[11]{0.0409, 0.0036, 0.0626, 0.0756, 0.0964, 0.0403, 0.0641, 0.0403, 0.0617, 0.1311, 0.0627} \\
                    \cline{2-6}
                    & \cellcolor{lightgray} Manual & \multicolumn{4}{c|}{\cellcolor{lightgray} \bf Manually Developed \& Optimized Model: 0.0058} \\
                    \cline{2-6}
                    \hline
            
                \multirow{10}{*}{\textbf{DJI}} 
                    & 1 & 0.0383 & NA & 0.0571 & 0.0705 \\
                    \cline{2-6}
                    & 2  & 0.0422 & NA & 0.1304 & 0.1021  \\
                    \cline{2-6}
                    & 3  & 0.0996 & \cellcolor{lightgray}{\bf 0.0550} & 0.0660 & 0.0602 \\
                    \cline{2-6}
                    & 4  & 0.0416 & NA & \cellcolor{lightgray}{\bf 0.0423} & \cellcolor{black}\textcolor{white}{\bf0.0034}  \\
                    \cline{2-6}
                    & 5  &\cellcolor{lightgray}{\bf 0.0373} & NA & 0.0494 & 0.1270 \\
                    \cline{2-6}
                    & 6  & 0.0418 & NA & 0.0462 & 0.0547\\
                    \cline{2-6}
                    & 7  & 0.0402 & 0.0984 & 0.0521 & 0.0462 \\
                    \cline{2-6}
                    & 8  & 0.0382 & NA & 0.0542 & 0.0531 \\
                    \cline{2-6}
                    & 9  & 0.0402 & NA & 0.0470 & 0.0631 \\
                    \cline{2-6}
                    & 10  & 0.0404 & NA & 0.0687 & 0.0490  \\
                    \cline{2-6}
                    & 11  & 0.0707 & NA & 0.0514 & 0.0493 \\
                    \cline{2-6}
                     & Avg. 
                     & \bf \avercalc[11]{0.0383, 0.0422, 0.0996, 0.0416, 0.0373, 0.0418, 0.0402, 0.0382, 0.0402, 0.0404, 0.0707} 
                     & \bf \avercalc[2]{0.0550, 0.0984} 
                     & \bf \avercalc[11]{0.0571, 0.1304, 0.0660, 0.0423, 0.0494, 0.0462, 0.0521, 0.0542, 0.0470, 0.0687, 0.0514} 
                     & \bf \avercalc[11]{0.0705, 0.1021, 0.0602, 0.0034, 0.1270, 0.0547, 0.0462, 0.0531, 0.0631, 0.0490, 0.0493}  \\
                    \cline{2-6}
                    & \cellcolor{lightgray} Manual & \multicolumn{4}{c|}{\cellcolor{lightgray} \bf Manually Developed \& Optimized Model: 0.0053} \\
                    \cline{2-6}
                    \hline
                
                \multirow{10}{*}{\textbf{IXIC}} 
                & 1 & \cellcolor{lightgray}{\bf 0.0259} & NA & 0.0372 & 0.1052 \\
                \cline{2-6}
                & 2  & 0.0299 & NA & 0.0324 & 0.0739  \\
                \cline{2-6}
                & 3  & 0.0300 & NA & \cellcolor{lightgray}{\bf 0.0318} & \cellcolor{black}\textcolor{white}{\bf0.0059} \\
                \cline{2-6}
                & 4  & 0.0307 & 0.1158 & 0.0327 & 0.0829  \\
                \cline{2-6}
                & 5  & 0.1676 & \cellcolor{lightgray}{\bf 0.0530} & 0.0339 & 0.1044 \\
                \cline{2-6}
                & 6  & 0.0298 & NA & 0.0328 & 0.0372\\
                \cline{2-6}
                & 7  & 0.0397 & NA & 0.0305 & 0.0697 \\
                \cline{2-6}
                & 8  & 0.0281 & NA & 0.0360 & 0.1886 \\
                \cline{2-6}
                & 9  & 0.0394 & NA & 0.0344 & 0.0345 \\
                \cline{2-6}
                & 10  & 0.0295 & NA & 0.0344 & 0.0323  \\
                \cline{2-6}
                & 11  & 0.0405 & NA & 0.0470 & 0.1280 \\
                \cline{2-6}
                 & Avg. 
                 & \bf \avercalc[11]{0.0259, 0.0299, 0.0300, 0.0307, 0.1676, 0.0298, 0.0397, 0.0281, 0.0394, 0.0295, 0.0405} 
                 & \bf \avercalc[2]{0.1158, 0.0530} 
                 & \bf \avercalc[11]{0.0372, 0.0324, 0.0318, 0.0327, 0.0339, 0.0328, 0.0305, 0.0360, 0.0344, 0.0344, 0.0470} 
                 & \bf \avercalc[11]{0.1052, 0.0739, 0.0059, 0.0829, 0.1044, 0.0372, 0.0697, 0.1886, 0.0345, 0.0323, 0.1280} \\
                    \cline{2-6}
                    & \cellcolor{lightgray} Manual & \multicolumn{4}{c|}{\cellcolor{lightgray} \bf Manually Developed \& Optimized Model: 0.0070} \\
                    \cline{2-6}
                    \hline
                
                \multirow{10}{*}{\textbf{N225}} 
                & 1 & 0.0275 & NA & 0.0511 & 0.0071 \\
                \cline{2-6}
                & 2  & 0.0307 & NA & 0.0398 & 0.0365  \\
                \cline{2-6}
                & 3  & 0.0311 & NA & 0.0636 & 0.1199 \\
                \cline{2-6}
                & 4  & 0.0257 & NA & 0.0391 & \cellcolor{black}\textcolor{white}{\bf0.0038}  \\
                \cline{2-6}
                & 5  & 0.0305 & NA & 0.0657 & 0.0410 \\
                \cline{2-6}
                & 6  & 0.0321 & NA & 0.0419 & 0.2064\\
                \cline{2-6}
                & 7  & 0.0270 & NA & \cellcolor{lightgray}{\bf 0.0268} & 0.0312 \\
                \cline{2-6}
                & 8  & 0.0313 & NA & 0.0552 & 0.1309 \\
                \cline{2-6}
                & 9  & 0.0368 & NA & 0.0599 & 0.0334 \\
                \cline{2-6}
                & 10  & 0.0326 & NA & 0.0363 & 0.0311  \\
                \cline{2-6}
                & 11  & \cellcolor{lightgray}{\bf 0.0211} & NA & 0.0582 & 0.1748 \\
                \cline{2-6}
                 & Avg. 
                & \bf \avercalc[11]{0.0275, 0.0307, 0.0311, 0.0257, 0.0305, 0.0321, 0.0270, 0.0313, 0.0368, 0.0326, 0.0211} 
                 & \bf NA 
                 & \bf \avercalc[11]{0.0511, 0.0398, 0.0636, 0.0391, 0.0657, 0.0419, 0.0268, 0.0552, 0.0599, 0.0363, 0.0582} 
                 & \bf \avercalc[11]{0.0071, 0.0365, 0.1199, 0.0038, 0.0410, 0.2064, 0.0312, 0.1309, 0.0334, 0.0311, 0.1748} \\
                    \cline{2-6}
                    & \cellcolor{lightgray} Manual & \multicolumn{4}{c|}{\cellcolor{lightgray} \bf Manually Developed \& Optimized Model: 0.0034} \\
                    \cline{2-6}
                    \hline
    
                \multirow{10}{*}{\textbf{HSI}} 
                & 1 & 0.0264 & NA & 0.0265 & 0.1928 \\
                \cline{2-6}
                & 2  & 0.0230 & 0.0225 & 0.0343 & 0.0314  \\
                \cline{2-6}
                & 3  & 0.1250 & NA & 0.0599 & 0.0428 \\
                \cline{2-6}
                & 4  & 0.0272 & NA & 0.0186 & \cellcolor{black} \textcolor{white} {\bf0.0068}  \\
                \cline{2-6}
                & 5  & 0.2486 & NA & 0.0500 & 0.1063 \\
                \cline{2-6}
                & 6  & 0.0278 & \cellcolor{lightgray}{\bf 0.0196} & 0.0240 & 0.0398\\
                \cline{2-6}
                & 7  & 0.0306 & NA & 0.0485 & 0.0367 \\
                \cline{2-6}
                & 8  & 0.0273 & NA & \cellcolor{lightgray}{\bf 0.0184} & 0.0268 \\
                \cline{2-6}
                & 9  & 0.0309 & NA & 0.0205 & 0.0622 \\
                \cline{2-6}
                & 10  & \cellcolor{lightgray}{\bf 0.0217} & NA & 0.0248 & 0.2108  \\
                \cline{2-6}
                & 11  & 0.0302 & 0.1508 & 0.0354 & 0.0434 \\
                \cline{2-6}
                 & Avg. 
                & \bf \avercalc[11]{0.0264, 0.0230, 0.1250, 0.0272, 0.2486, 0.0278, 0.0306, 0.0273, 0.0309, 0.0217, 0.0302} 
                 & \bf \avercalc[3]{0.0225,0.0196,0.1508} 
                 & \bf \avercalc[11]{0.0265, 0.0343, 0.0599, 0.0186, 0.0500, 0.0240, 0.0485, 0.0184, 0.0205, 0.0248, 0.0354} 
                 & \bf \avercalc[11]{0.1928, 0.0314, 0.0428, 0.0068, 0.1063, 0.0398, 0.0367, 0.0268, 0.0622, 0.2108, 0.0434} \\
                    \cline{2-6}
                    & \cellcolor{lightgray} Manual & \multicolumn{4}{c|}{\cellcolor{lightgray} \bf Manually Developed \& Optimized Model: 0.0354} \\
                    \cline{2-6}
                    \hline
              \end{tabular}
          }
    \end{subtable}%
    \begin{subtable}{0.5\linewidth}
      \centering
    \scalebox{0.75}{
    \begin{tabular}{|c|c|c|c|c|c|}
      \hline
      \multirow{2}{*}{\textbf{Ticker}} & \multirow{2}{*}{\textbf{Prompts}} & \multicolumn{4}{c|}{\textbf{RMSE}} \\
     \cline{3-6}
        &  & \textbf{PaLM} & \textbf{falcon} & \textbf{LLama 2} & \textbf{GPT 3.5} \\
      \hline
  
\multirow{10}{*}{\textbf{AAPL}} 
        & 1 & 0.0366 & NA & 0.0512  & 0.0564  \\
        \cline{2-6}
        & 2  & 0.0380 & NA & 0.0589  & 0.1250   \\
        \cline{2-6}
        & 3  & 0.1288 & NA & 0.0565  & 0.1144  \\
        \cline{2-6}
        & 4  & 0.0371 & 0.0388 & \cellcolor{lightgray}{\bf 0.0396}  & \cellcolor{black}\textcolor{white}{\bf 0.0029}  \\
        \cline{2-6}
        & 5  & 0.0411 & NA & 0.0412  & 0.4194   \\
        \cline{2-6}
        & 6  & 0.0366 & NA & 0.0458  & 0.4048  \\
        \cline{2-6}
        & 7  & 0.0358 & NA & 0.0414  & 0.4194  \\
        \cline{2-6}
        & 8  & 0.0359 & NA & 0.0428  & 0.1224   \\
        \cline{2-6}
        & 9  & \cellcolor{lightgray}{\bf 0.0360} & \cellcolor{lightgray}{\bf 0.0368} & 0.0448  & 0.4194  \\
        \cline{2-6}
        & 10  & \cellcolor{lightgray}{\bf 0.0360} & NA & 0.0457  & 0.0050   \\
        \cline{2-6}
        & 11  & 0.0361 & NA & 0.0522  & 0.4194   \\
        \cline{2-6}
         & Avg. 
         & \bf \avercalc[11]{0.0366, 0.0380, 0.1288, 0.0371, 0.0411, 0.0366, 0.0358, 0.0359, 0.0360, 0.0360, 0.0361} 
         & \bf \avercalc[2]{0.0388,0.0368} 
         & \bf \avercalc[11]{0.0512, 0.0589, 0.0565, 0.0396, 0.0412, 0.0458, 0.0414, 0.0428, 0.0448, 0.0457, 0.0522} 
         & \bf \avercalc[11]{0.0564, 0.1250, 0.1144, 0.0029, 0.4194, 0.4048, 0.4194, 0.1224, 0.4194, 0.0050, 0.4194} \\
                    \cline{2-6}
        & \cellcolor{lightgray} Manual & \multicolumn{4}{c|}{\cellcolor{lightgray} \bf Manually Developed \& Optimized Model: 0.0094} \\
        \cline{2-6}
        \hline
\multirow{10}{*}{\textbf{MSFT}} 
        & 1 & 0.0399 & NA & 0.0468  & 0.2605  \\
        \cline{2-6}
        & 2  & 0.0367 & NA & 0.0428  & 0.1850   \\
        \cline{2-6}
        & 3  & 0.1264 & NA & 0.0444  & \cellcolor{lightgray}{\bf 0.0450}  \\
        \cline{2-6}
        & 4  & \cellcolor{black}\textcolor{white}{\bf 0.0357} & \cellcolor{lightgray}{\bf 0.0591} & 0.0461  & 0.2456  \\
        \cline{2-6}
        & 5  & 0.0363 & NA & 0.0436  & 0.1241   \\
        \cline{2-6}
        & 6  & 0.0407 & NA & 0.0482  & 0.2557  \\
        \cline{2-6}
        & 7  & 0.0392 & NA & \cellcolor{lightgray}{\bf 0.0419}  & 0.1270  \\
        \cline{2-6}
        & 8  & 0.0401 & NA & 0.0495  & 0.2060   \\
        \cline{2-6}
        & 9  & 0.0418 & NA & 0.0475  & 0.1937  \\
        \cline{2-6}
        & 10  & 0.0392 & NA & 0.0469  & 0.0478   \\
        \cline{2-6}
        & 11  & 0.0390 & NA & 0.0460  & 0.1581   \\
        \cline{2-6}
         & Avg. 
         & \bf \avercalc[11]{0.0399, 0.0367, 0.1264, 0.0357, 0.0363, 0.0407, 0.0392, 0.0401, 0.0418, 0.0392, 0.0390} 
         & \bf \avercalc[1]{0.0591} 
         & \bf \avercalc[11]{0.0468, 0.0428, 0.0444, 0.0461, 0.0436, 0.0482, 0.0419, 0.0495, 0.0475, 0.0469, 0.0460} 
         & \bf \avercalc[11]{0.2605, 0.1850, 0.0450, 0.2456, 0.1241, 0.2557, 0.1270, 0.2060, 0.1937, 0.0478, 0.1581} \\
                    \cline{2-6}
        & \cellcolor{lightgray} Manual & \multicolumn{4}{c|}{\cellcolor{lightgray} \bf Manually Developed \& Optimized Model: 0.0065} \\
        \cline{2-6}
        \hline
\multirow{10}{*}{\textbf{AMZN}} 
        & 1 & \cellcolor{lightgray}{\bf 0.0414} & \cellcolor{lightgray}{\bf 0.0399} & \cellcolor{lightgray}{\bf 0.0438}  & 0.2763  \\
        \cline{2-6}
        & 2  & 0.0422 & NA & 0.0494  & 0.4927   \\
        \cline{2-6}
        & 3  & 0.1270 & NA & 0.0472  & 0.0046  \\
        \cline{2-6}
        & 4  & 0.0417 & 0.0657 & 0.0477  & \cellcolor{black}\textcolor{white}{\bf 0.0043}  \\
        \cline{2-6}
        & 5  & 0.1017 & NA & 0.0455  & 0.0866   \\
        \cline{2-6}
        & 6  & 0.0420 & NA & 0.0487  & 0.0545  \\
        \cline{2-6}
        & 7  & 0.0426 & NA & 0.0462  & 0.4144  \\
        \cline{2-6}
        & 8  & 0.0426 & NA & 0.0485  & 0.1512   \\
        \cline{2-6}
        & 9  & 0.0425 & NA & 0.0470  & 0.4655  \\
        \cline{2-6}
        & 10  & 0.0430 & NA & 0.0456  & 0.0054   \\
        \cline{2-6}
        & 11  & 0.0418 & NA & 0.0461  & 0.4771   \\
        \cline{2-6}
         & Avg. 
         & \bf \avercalc[11]{0.0414, 0.0422, 0.1270, 0.0417, 0.1017, 0.0420, 0.0426, 0.0426, 0.0425, 0.0430, 0.0418} 
         & \bf \avercalc[2]{0.0399,0.0657} 
         & \bf \avercalc[11]{0.0438, 0.0494, 0.0472, 0.0477, 0.0455, 0.0487, 0.0462, 0.0485, 0.0470, 0.0456, 0.0461} 
         & \bf \avercalc[11]{0.2763, 0.4927, 0.0046, 0.0043, 0.0866, 0.0545, 0.4144, 0.1512, 0.4655, 0.0054, 0.4771} \\
                    \cline{2-6}
        & \cellcolor{lightgray} Manual & \multicolumn{4}{c|}{\cellcolor{lightgray} \bf Manually Developed \& Optimized Model: 0.0062} \\

        \cline{2-6}
        \hline
\multirow{10}{*}{\textbf{BABA}} 
        & 1 & 0.0274 & NA & 0.0258  & 0.0292  \\
        \cline{2-6}
        & 2  & \cellcolor{lightgray}{\bf 0.0229} & NA & 0.0299  & 0.0668   \\
        \cline{2-6}
        & 3  & 0.1577 & NA & 0.0291  & \cellcolor{black}\textcolor{white}{\bf0.0211}  \\
        \cline{2-6}
        & 4  & 0.0247 & NA & 0.0306  & 0.0295  \\
        \cline{2-6}
        & 5  & 0.0780 & 0.1597 & 0.0279  & 0.0738   \\
        \cline{2-6}
        & 6  & \cellcolor{lightgray}{\bf 0.0229} & NA & 0.0278  & 0.0444  \\
        \cline{2-6}
        & 7  & 0.0245 & \cellcolor{lightgray}{\bf 0.0225} & 0.0287  & 0.1610  \\
        \cline{2-6}
        & 8  & 0.0232 & NA & \cellcolor{lightgray}{\bf 0.0220}  & 0.2473   \\
        \cline{2-6}
        & 9  & 0.0252 & NA & 0.0248  & 0.2474  \\
        \cline{2-6}
        & 10  & 0.0243 & NA & 0.0296  & 0.2042   \\
        \cline{2-6}
        & 11  & 0.0234 & NA & 0.0879  & 0.1755   \\
        \cline{2-6}
         & Avg.  
         & \bf \avercalc[11]{0.0274, 0.0229, 0.1577, 0.0247, 0.0780, 0.0229, 0.0245, 0.0232, 0.0252, 0.0243, 0.0234} 
         & \bf \avercalc[2]{0.1597, 0.0225} 
         & \bf \avercalc[11]{0.0258, 0.0299, 0.0291, 0.0306, 0.0279, 0.0278, 0.0287, 0.0220, 0.0248, 0.0296, 0.0879} 
         & \bf \avercalc[11]{0.0292, 0.0668, 0.0211, 0.0295, 0.0738, 0.0444, 0.1610, 0.2473, 0.2474, 0.2042, 0.1755} \\
                    \cline{2-6}
        & \cellcolor{lightgray} Manual & \multicolumn{4}{c|}{\cellcolor{lightgray} \bf Manually Developed \& Optimized Model: 0.0268} \\

        \cline{2-6}
        \hline
\multirow{10}{*}{\textbf{TSLA}} 
        & 1 & \cellcolor{lightgray}{\bf 0.0325} & NA & 0.0430  & \cellcolor{black}\textcolor{white}{\bf 0.0024}  \\
        \cline{2-6}
        & 2  & 0.0349 & 0.1266 & 0.0484  & 0.0423   \\
        \cline{2-6}
        & 3  & 0.0695 & NA & 0.0603  & 0.0968  \\
        \cline{2-6}
        & 4  & 0.0348 & NA & 0.0445  & 0.0076  \\
        \cline{2-6}
        & 5  & NA & NA & 0.0402  & 0.0630   \\
        \cline{2-6}
        & 6  & 0.0330 & \cellcolor{lightgray}{\bf 0.0327} & \cellcolor{lightgray}{\bf 0.0375}  & \cellcolor{black}\textcolor{white}{\bf 0.0024}  \\
        \cline{2-6}
        & 7  & 0.0369 & NA & 0.0565  & 0.0052  \\
        \cline{2-6}
        & 8  & 0.0364 & 0.5569 & 0.0405  & 0.0432   \\
        \cline{2-6}
        & 9  & 0.0353 & NA & 0.0525  & 0.1006  \\
        \cline{2-6}
        & 10  & 0.0353 & 0.5663 & 0.0414  & NA   \\
        \cline{2-6}
        & 11  & 0.0379 & NA & 0.0377  & 0.4147   \\
        \cline{2-6}
         & Avg.  
         & \bf \avercalc[10]{0.0325, 0.0349, 0.0695, 0.0348, 0.0330, 0.0369, 0.0364, 0.0353, 0.0353, 0.0379} 
         & \bf \avercalc[4]{ 0.1266,0.0327,0.5569,0.5663} 
         & \bf \avercalc[11]{0.0430, 0.0484, 0.0603, 0.0445, 0.0402, 0.0375, 0.0565, 0.0405, 0.0525, 0.0414, 0.0377} 
         & \bf \avercalc[10]{0.0024, 0.0423, 0.0968, 0.0076, 0.0630, 0.0024, 0.0052, 0.0432, 0.1006, 0.4147} \\
                    \cline{2-6}
        & \cellcolor{lightgray} Manual & \multicolumn{4}{c|}{\cellcolor{lightgray} \bf Manually Developed \& Optimized Model: 0.0100} \\
        \cline{2-6}
        \hline 
    \end{tabular}
}
    \end{subtable} 
\end{table*}

As Table \ref{tab:result2} indicates that the reduction in randomness through minimizing the value of the temperature value has some impacts on the performance of each prompt. The table demonstrates that the GPT 3.5 model achieved the lowest RMESE for nine tickers whereas Palm achieved the lowest RMSE value of 0.0357 for the MSFT ticker.

{\it I) The Performance of Generated Models Across LLMs.} A detailed view on both Tables \ref{tab:result1} and \ref{tab:result2} indicates that lower values of temperature makes the accuracy of models slightly better. In particular, we observe that GPT still outperforms other LLMs. 

{\it II) The Performance of Generated Models Across Prompts.} We observe that the the best models generated by GPT are the ones generated by simpler prompts such as Prompt 2, 3, and 4 where the criteria (i.e., Clarity and Specificity, Objective and Intent, Contextual Information, and Format and Style) are all kept consistent at the level of either low or medium or high. 

{\it III) The Performance of Generated Models Across the Time Series Datasets.} A similar pattern is observed. A mixed results, but consistent with the results observed in Table \ref{tab:result1}.

{\it IV) The Performance of Generated Models and Manually Developed and Optimized Model.} As shown in both Tables \ref{tab:result1} and \ref{tab:result2}, we observe a slightly better models for the case where the temperature parameter is kept low. 

Tables \ref{tab:result1} and \ref{tab:result2} clearly demonstrated that the Falcon model generates more {\it valid} and {\it correct} models  when the temperature parameter is configured at 0.7 (high) compared to 0.1 (low). The results show the number of invalid models labeled with ``NA'' is lower than the number of invalid models generated by higher temperature 0.7 which leads to more exploration in the model's predictions. By increasing the temperature, the model is encouraged to introduce more randomness into its predictions, reducing the likelihood of exhibiting the hallucinated phenomenon.

In contrast, for GPT 3.5 Turbo model the number of invalid models (i.e., "NA") is lower with temperature parameter set to low (i.e., $0.1$) instead of high (i.e., $0.7$). The simpler prompts with lower temperature yield better results because the model produce more coherent and relevant responses. In case of complex prompts with higher temperature the GPT 3.5 model explores a wider range of possibilities and generate more diverse responses because of higher randomness. Complex prompts may contain unclear information, making it difficult for the model to provide appropriate outputs with high confidence. In such circumstances, greater temperatures allow the model to experiment with different variations of the prompt, resulting in responses that represent the input's nuances.


\subsection{Model Architecture of Generated Models}

Given the variation in performance of the models generated by LLMs, it is of important to investigate the cause of such differences. One of the key factors in deep learning-based models including LSTM, which plays an important role in the performance, is the architecture (e.g., number of layers and nodes) of the generated models. To compare the architecture of the models generated by LLMs and the architecture of our manually created and optimized LSTM model, this section reports the architecture metadata of all models.

Table \ref{tab:arch1} reports configuration of the models generated by LLMs using the prompts listed in Table  \ref{ref:prompts_1}. The configurations are set differently for each LSTM model with a number of hyperparameters to analyze. The configuration consists of 1) the number of the number of LSTM layers, 2) number of units, 3) activation function, 4) batch sizes, and 5) epochs. 

In Table 6, we see a summary of LSTM model architecture configurations by different LLMs with their different prompts. It also specifies such parameters as the number of LSTM layers, the number of units per layer to use, and what activation functions to use, as well as batch sizes and number of epochs used. These configurations are crucial, as few layers and units tend to lead to more capacity, but on the flip side are more likely to overfit while more layers and units give more capacity. As learning dynamics depend on the choice of activation function, batch size and epoch, which determines training stability and efficiency. The large variation in these architectural choices emphasizes the role that prompt design plays in determining model performance as measured by corresponding RMSE values.

The architecture of the manually created and optimized model is configured as: one LSTM layer of 50 units (i.e., nodes), where ``relu'' is used as the activation function, with the batch size and epochs set to 1 and 100, respectively. The manually created and optimized model is based on all data studied in this work implying that only one model manually created and optimized to represent the entire datasets.


\begin{table*}
\footnotesize
    \centering
    \caption{{Model Architecture Details: P. = Prompt; Format=[LSTM Layer, Units, Activation, Batch, Epoch];  \\ Manually Created and Optimized Model: [1, 50, 'relu', 1, 100].\\ 
    }}
    \label{tab:arch1}
\vspace{-0.3cm}
    \begin{subtable}{0.5\linewidth}
      \centering
       \scalebox{0.5}{
        \begin{tabular}{|c|c|c|c|c|c|}
          \hline
          \textbf{Ticker} &\textbf{P.}  & \textbf{PaLM} & \textbf{falcon} & \textbf{LLama 2} & \textbf{GPT 3.5} \\
          \hline
      
                 \multirow{10}{*}{\textbf{GSPC}} 
                    & 1 & [1, 128, NA, 32, 100],  & [1,64,NA, 16, 100] & [1, 50, NA, 32, 100] & NA  \\
                    
                    & 2 & [1,50, 'relu',32,100] & [1,128,NA, 128,1] & [2, 50, NA, 32, 100] &[1, 50, Na, 32, 50]  \\
                    
                    & 3 & [1,128,NA, NA, 100]  & [3, 128, NA, 256, 100] & [2, [50,64], 'relu', 32, 100]& NA  \\
                    
                    & 4 & [1,100,'relu',32,100]  & [1, 64, NA, 128,10] & [1, 50, 'relu', 32, 100]& NA  \\
                    
                    & 5 & [1,128,'relu',32,100] &NA & [2, [50,32], NA, 32, 100] & [1, 50, NA, 32, 10] \\
                    
                    & 6 & [1,100,'relu',16,100],  & [2, 128, NA, 128, 128] & [1, 50, NA, NA, 100]& NA  \\
                    
                    & 7 & [1,100,'relu',20,100],  & [2, 128, NA, 32, 500] & [2, 50, NA, NA, 100]& NA  \\
                    
                    & 8 & [1,128, NA,32,100] & [1, 128, NA, NA, NA] & [1, 128, NA, NA, 100]&[1,50,'relu',32, 10]  \\
                    
                    & 9 & [1,100,'relu',32,100] &NA & [2, [50,32], NA, 32, 100] & [1,50,'relu',32, 50]  \\
                    
                    & 10 & [1,128,NA,32,100] & [1, 128, NA, 128, 100] & [1, 50, NA, 32, 100] & [1,50,NA,1, 100] \\
                    
                    & 11 & [1,100,'relu',NA,100] & [1, 10, NA, 256, 100]& [2, [50,32], NA, 32, 50] & NA  \\
                    
                    \cline{2-6}
                    \hline
            
                \multirow{10}{*}{\textbf{DJI}} 
                    & 1 & [1,100, 'relu',32,100] & NA & [1,50, NA,32,100]& [2,[50,50], NA,1,100]\\
                   
                    & 2 & [1,100, NA,32,100] & NA &[1,50,'relu',32,100] & [1,50,'relu',32,10]\\
                   
                    & 3 & [1,128, NA,32,100] & [1, 128, NA, 32,NA] & [2,[50,64],NA,32,100] & [1,50, NA,32,50] \\
                    
                    & 4 & [1,100, NA,16,100] & [1, 128, NA, NA, 100] & [1,50, NA,NA,100] & [1,50, NA,32,10]  \\
                    
                    & 5 & [1,128, 'relu',32,100]& NA & [2,[50,32],NA,32,100] & [1,64, NA,32,10]  \\
                    
                    & 6 & [1,100, 'relu',30,100] & NA & [1,50, NA,NA,100] & NA \\
                    
                    & 7 & [1,100, 'relu',NA,100] & NA & [2,50, NA,32,100] & NA \\
                    
                    & 8 & [1,128, NA,32,100] & [1, 256, NA, 256, NA] & [1,128, NA,NA,100] & [1,50, 'relu',NA,10] \\
                    
                    & 9 & [1,128,'relu',32,100] & NA & [2,[50,32], NA,NA,100] & [2,[50,50], 'relu',1,100] \\
                    
                    & 10 & [2,[128,64], NA,32,100] & NA & [1,50, NA,NA,100] & [1,50, NA,16,10]  \\
                    
                    & 11 & [1,128, NA,NA,100] & NA & [2,[50,32], NA,32,50] & NA  \\
                    
                    \hline
                
                \multirow{10}{*}{\textbf{IXIC}} 
                   & 1 & [1,128,'relu',32,100] & NA &[1,50, NA,32,100] & [2,[50,50], NA,32,10]  \\
                    
                    & 2 & [1,128, NA,32,100] & NA & 1,50, NA,NA,100 & NA  \\
                    
                    & 3 & 1,128, NA,NA,100 & NA & [2,[50,64], NA,32,100] & NA \\
                    
                    & 4 & [1,100,'relu',32,100] & [1,100, NA, 256, 100] & [1,50, 'relu',32,100] & NA  \\
                    
                    & 5 & [1,128, NA,NA,100] & [1,100,NA, 32, 10] & [2,[50,32], NA,32,100] & [1,64, NA,32,10]  \\
                   
                    & 6 & [1,128, NA,NA,NA] & [3,[128,10,10],NA,256,100] & [1,50, NA,NA,100] & NA \\
                   
                    & 7 & [1,100, NA,32,100] & NA & [1,50,'relu',NA,100] & NA \\
                    
                    & 8 & [1,50, NA,32,100] & [1,128,NA,10,NA] & [1,128, NA,NA,100] & [2,[50,50],NA,32,100]  \\
                    
                    & 9 & [1,50,'relu',32,100] & [4,[100,200,300,400] NA, NA,NA] & [2,[50,32],NA,NA,10] & NA \\
                    
                    & 10 & [1,50, NA,NA,100] & NA & [1,50,NA,NA,100] & [1,50, NA,1,100]  \\
                    
                    & 11 & 1,128, NA,16,100 & NA & [2,[50,32],NA,32,50] & NA  \\
                    
                    \hline
                
                \multirow{10}{*}{\textbf{N225}} 
                    & 1 & NA & [1,1,NA,32,100] & [1,50, NA,32,100] & NA  \\ 
                    
                    & 2 & [2,[128,64], NA,NA,100] & NA & [1,50, NA,1,100] & NA  \\
                    
                    & 3 & NA & NA & [2,[50,64],NA,NA,100] &[1,50, NA,32,100] \\
                
                    & 4 & [2,[128,64],'relu',32,100] &[1, 32, NA, 32,NA] & [1,50, 'relu',32,100] & NA  \\ 
                    
                    & 5 & [2,[128,64],'linear',32,100]& [3,[128,256,256],NA, NA, NA] & [2,[50,32],NA,32,100] & [2,[64,64], NA,32,10]  \\
                    
                    & 6 & [2,[128,64], NA,NA,100] & NA & [1,50, NA,NA,100] & [1,50,'relu',32,50] \\
                    
                    & 7 & [2,[128,64],'relu',NA,100] & NA &[1,50,'relu',NA,100] & NA \\
                    
                    & 8 & NA & NA & [1,128, NA,NA,100] & [2,[50,50],NA,32,100]  \\ 
                    
                    & 9 & [2,[128,64],'relu',16,100] & NA & [2,[50,32],'relu',NA,100] &[2,[50,50], NA,1,100]\\
                    
                    & 10 & [1,128,'relu',NA,100] & [1,2,NA, 32,1000] & [1,50, NA,32,50] & [1,50, NA,1,100]  \\
                    
                    & 11 &[1,128, NA,NA,100] & NA & [2,[50,32],NA,32,50] & NA  \\
                    
                    \hline
    
                \multirow{10}{*}{\textbf{HSI}} 
                    & 1 & NA & [1,128,NA,10,NA] & [1,50, NA,32,100] & [1,50, NA,1,100] \\ 
                    
                    & 2 & NA & [2,[64,128],NA, 32, NA] & [1,50,'relu',32,100] &[1,50, 'relu',1,100] \\
                    
                    & 3 & NA & NA &[2,[50,64],NA,32,100] & NA  \\ 
                    
                    & 4 & NA & NA & [1,50, NA,NA,100] & [1,50, NA,NA,100] \\
                    
                    & 5& NA & NA & [2,[50,32],NA,NA,100] & [1,50, NA,32,100] \\
                    
                    & 6 & NA & NA & [1,50, NA,NA,100] & [1,50, 'relu',NA,100] \\
                    
                    & 7 & NA &[1,10,NA,1,NA] & [1,50,'relu',NA,100] & [2,[50,50],NA,32,100] \\
                    
                    & 8 & [1,100,NA,32,100] & NA & [1,128, NA,32,100] & [2,[50,50],'relu',32,100]  \\
                    
                    & 9 & [2,100,NA,32,100] & NA & [2,[50,32],NA,NA,100] & [1,4, NA,1,100] \\
                    
                    & 10 & NA & NA & [2,50,NA,32,100] & [1,50, NA,32,100] \\
                    
                    & 11 & [1,50,NA,32,100] & NA & [2,[50,32],NA,32,50] & NA  \\
                    
                    \hline

        \end{tabular}
        }
      \end{subtable}%
    \begin{subtable}{0.5\linewidth}
      \centering
      \scalebox{0.5}{
    \begin{tabular}{|c|c|c|c|c|c|}
      \hline
      \textbf{Ticker} &\textbf{P.} &  \textbf{PaLM} & \textbf{falcon} & \textbf{LLama 2} & \textbf{GPT 3.5} \\
      \hline
  
\multirow{10}{*}{\textbf{AAPL}} 
        & 1 & [1,100,NA,32,100] & NA & [1,50, NA,32,100]  &[1,50,NA,NA,100] \\
        
        & 2 & [1,128, NA, NA, 100] & [1,128, NA, NA, NA]& [1,50,'relu',32,100]  & [2,[50,50], NA, 16, 100]  \\
        
        & 3 & [1,128, NA,32,10] & NA & [2,[50,64],NA,NA,100] & [1,50, NA,32,10]  \\
        
        & 4 & [1,100,'relu',32, 100] & NA& [1,50, NA,NA,100]  & [1,50, NA,32, 50] \\
        
        & 5 & [1,100, 'relu', 32, 10] & [1,32,NA, 10, NA]& [2,[50,32],NA,NA,100]  & [1,128, NA, 32, 10] \\
        
        & 6 & [1,100,NA,32,10]& NA& [1,50, NA,NA,100]  & [1,50,NA,32,100] \\
        
        & 7 & [1,100,NA,16,100] & [2,10,NA,32, NA]& [1,50, NA,1,100]  &  [1,50,NA,16,10]\\
        
        & 8 & [1,100,'relu',32,10] & NA& [1,128, NA,NA,100]  & [2,[50,50],NA,32,10]  \\
        
        & 9 & [1,128,'relu',32,100] & NA& [2,[50,32],NA,NA,100]   & [1,100,NA,32,10] \\
        
        & 10 &[1,100,NA,1,100] & [1,1,NA,32,NA]& [[1,50,'relu',1,100]]  & [2,[50,50],NA,1,100] \\
        
        & 11 & [1,100,'relu',1,10] & NA& [2,[50,32],NA,32,50] & [1,64,NA,32,10]  \\
        
        \hline
\multirow{10}{*}{\textbf{MSFT}} 
        & 1 & [1,100,'relu',32,100] & [1,128,NA, 128, 100] & [1,50, NA,NA,100] & [2,[50,50],NA,32,10] \\
        
        & 2 & [1,100,NA,NA,100] & NA & [1,50, NA,32,100] & [1,50, 'relu',16,10]  \\
        
        & 3 & [1,128,NA,32,10] & NA &[2,[50,64],NA,32,100] & [1,128,NA,32,10] \\
        
        & 4 & [1,100,NA,32,100] & [1,99,NA,1,100] & [1,50, 'relu',NA,100]& [1,50, 'relu',1,100] \\
        
        & 5 & [2,[128,128],NA,32,10] & NA & [2,[50,32],NA,32,100] & [1,64,NA,32,10]  \\
        
        & 6 & [1,50, 'relu',32,10] & [1,100,NA,100,NA] & [[1,50, NA,1,100]] & [1,50, 'relu',32,10] \\
        
        & 7 & [1,100, 'relu',NA,100] & NA & [1,50, 'relu',1,100] & [1,50, 'relu',NA,10] \\
        
        & 8 & [1,100,NA,32,10] & NA & [[1,128, 'relu',NA,100]] & [2,[50,50],NA,32,10]  \\
        
        & 9 & [1,128,NA,32,100]& [1,512,NA,1,NA] & [2,[50,32],NA,32,100] & [2,[50,50],'relu',32,10] \\
        
        & 10 & [1,128, 'relu',NA,100] & NA & [[1,50,NA,NA,10]] & [1,64, 'relu',32,10]   \\
        
        & 11 & [1,128, 'relu',32,100] & [1,64, NA, 32, NA] & [2,[50,32],NA,32,50] & [1,50, 'relu',32,100]   \\
        
        \hline
\multirow{10}{*}{\textbf{AMZN}} 
        & 1 & [1,100, NA,32,100]& [1,32,NA,256,NA] & [1,50,'relu',32,100] & [1,50, NA,32,50] \\
        
        & 2 & [1,100,NA,32,10] & [1,128,NA,32,NA] & [1,50,NA,32,100] & [2,[50,50],NA,32,10] \\
        
        & 3 & [1,128,NA,32,10] & NA & [2,[50,64],NA,32,100] & [1,50,NA,32,100] \\
        
        & 4 & [1,100, 'relu',32,100]& NA & [1,50,NA,16,100] &[1,50, 'relu',32,100]  \\
        
        & 5 & [1,128,'relu',32,10] & [1,32,NA,32,NA] & [2,[50,32],NA,32,100] & [1,128,NA,32,100]  \\
        
        & 6 & [1,100,'relu',32,10] & NA & [1,50,NA,NA,100] & [1,50, 'relu',32,10] \\
        
        & 7 & [1,100, 'relu', NA,100] & NA & [1,50,NA,1,100] & [1,50, 'relu', NA,100] \\
        
        & 8 & [1,128,NA,32,100] & NA & [1,128,NA,NA,100] & [1,50,NA,1,10]  \\
        
        & 9 & [1,100,NA,1,100] & NA & [2,[50,32],NA,NA,100] &[1,4,NA,1,100] \\
        
        & 10 & [1,100, 'relu',1,10] & NA & [1,50, NA, 16,100] & [1,50, 'relu',16,10] \\
        
        & 11 & [1,100, NA,16,10] & NA & [1,[50,32],NA,32,50] & [1,50, NA,16,100] \\

        \hline
\multirow{10}{*}{\textbf{BABA}} 
        & 1 & [1,100,'relu',32,100] & [1,16,NA,32,NA] &[1,50,NA,32,100] & [2,[50,50],NA,32,10] \\
        
        & 2 & [1,100, 'relu',32,10] & NA & [1,50, 'relu',NA,100] & [1,50, 'relu',32,10]  \\
        
        & 3 & [1,128,NA,32,10] & NA & [2,[50,64],NA,32,100] & [1,50,NA,32,10]  \\
        
        & 4 & [1,100, 'relu',NA,100] & NA & [1,50, 'relu',32,100] & [1,50, 'relu',32,100]  \\
        
        & 5 & [1,128, 'relu',32,10] & [1,10,NA,10,10] &[2,[50,32],NA,32,100] & [1,128, 'relu',32,10]  \\
       
        & 6 &[1,64, NA,32,10] & [1,128,NA,1,NA] & [1,50, 'relu',1,100] & [1,50, 'relu',NA,10] \\
        
        & 7 &[1,100, NA,1,10] & [1,32,NA,128,NA] & [1,50,NA,1,100] & [1,50, NA,16,10] \\
        
        & 8 & [1,128,NA,32,100] & NA & [1,128,NA,NA,100] & [2,[50,50],NA,32,10]  \\
        
        & 9 & [1,100,NA,NA,100] & NA & [2,[50,32],NA,100] & NA \\
        
        & 10 & [1,100,NA,NA,10] & [1,64,NA,64,1000] &[1,50, 'relu',16,100] & [2,[50,50],NA,32,100]  \\
        
        & 11 & [1,128,NA,NA,100] & NA & [2,[50,32],NA,32,50] & [1,64,NA,32,10]  \\
        
        
        \hline
\multirow{10}{*}{\textbf{TSLA}} 
        & 1 & [1,100,NA,32,100] & NA & [1,50, 'relu',32,100] & [2,[50,50],NA,32,10]  \\
        
        & 2 & [1,100,'relu',32,100] & [1,128,NA,64,NA] & [1,50, 'relu',NA,100] & [2,[50,50],'relu',32,10]  \\
        
        & 3 & [2,[128,10],NA,NA,10] & NA & [2,[50,64],NA,32,100] & [1,50,NA,1,100] \\
        
        & 4 & [1,100,'relu',NA,100] & [1,128,NA,256,100] & [1,50,NA,1,100] & [1,50,'relu',1,100]  \\
        
        & 5 & [1,128,'relu',32,100] & NA & [2,[50,32],NA,32,100] & [1,64,NA,1,100] \\
        
        & 6 & [1,100,'relu',32,10] & [1,128, NA, 128, 100] & [1,50,NA,NA,100] & [1,50,'relu',32,10] \\
        
        & 7 & [1,100,NA,NA,100] & NA & [1,50,NA,16,100] & [1,50,NA,1,10] \\
        
        & 8 & [1,100,NA,32,10] & [2,[32,32],NA,NA,NA] & [1,128,NA,NA,100] & [2,[50,50],NA,16,10]  \\
        
        & 9 & [1,128,NA,NA,100] & [1,32,NA,32,100] & [2,[50,32],NA,NA,100] & [1,50,NA,16,10] \\
        
        & 10 & [1,100,NA,16,100] &[1,256,NA,NA,NA] & [1,50, 'relu',16,100] & [1,50,NA,32,100]  \\
        
        & 11 & [1,100,'relu',16,100] & NA & [2,[50,32],NA,32,50] & [2,[50,50],NA,32,100]  \\
        
        \hline

    \end{tabular}
    }
    \end{subtable}%
        
\end{table*}

As Table \ref{tab:arch1} indicates, in most cases, the models generated by LLMs contain 1 or 2 LSTM layer (i.e., the first component in the architecture model), which is relatively consistent with the architecture of the manual model, where the number of LSTM layer is set to 1. 

The key difference between the models generated by LLMs and the manual model architecture is the number of nodes (i.e., unit), or the second component in the architecture. The LSTM-based models generated by PaLM and falcon consider a large number for the number of units or nodes in their LSTM model (e.g., 128, 100, 64). On the other hand, the number of units or nodes in the manually generated model is set to 50. A quick inspection of the number of nodes considered for LLama 2 and GPT 3.5 indicates that these two LLMs have considered the number of nodes as 50, which is similar to the number of nodes set for the manual model. This observation may explain the better performance and accuracy obtained by the LSTM models generated by LLama 2 and GPT 3.5 compared to PaLM and falcon. 

The employed activation function in most cases is either `relu' or NA. As a result this parameter of the architecture cannot be considered for comparison purposes. On the other hand, the batch size parameter is where we observe some ``additional' improvement are achieved. Most of batch sizes set by LLama 2 and GPT 3.5 (the two outperforming LLMs in generating better models) have set their batch size to 32; whereas, the batch size in our manually created and optimized model is set to 1. From the literature, we know that the smaller value of batch size helps in training models more profoundly.
The epoch parameter seems set mostly to 100, 50, 10 by all LLMs. In particular, the value of epochs is set to 100 in all instance models generated by PaLM without any variations. 

The take away lessons are : 
\begin{enumerate}
\item GPT 3.5 generates LSTM-based models with model architecture that are relatively similar to the architecture of the manually created and optimized model. GPT 3.5 is followed by LLama 2 in generating the most similar architectures for the models. On the other hand, the architecture of the models generated by PaLM and falcon are less similar to the manually created and optimized model by an expert. 
\item Most LLMs generate deep learning LSTM-based models with number of layers equal to 1 or in some rare cases to 2, which is consistent with the architecture of our manually generated model. 
\item The key parameter in architecture, that seems to contribute significantly to the accuracy of the models generated, is the number of nodes or units considered for each model. While PaLM and falcon consider a large value for hte number of units or nodes (e.g., 128), the models generated by GPT 3.5 and LLama 2 consider similar number of nodes and units (i.e., 50) compare to the manually created models. This observation indicates that the number of nodes plays a key role in improving the accuracy of the models generated, and GPT 3.5 and LLama 2 set this parameter better than the other two LLMs namely PaLM and falcon.
\item The second key contributor to the accuracy of models seems to be the value of batch size. While the two most outperforming LLMs in generating better model set their batch size to 32, other manually created and optimized model sets the batch size to 1, capturing additional and in-depth patterns in the data. An interesting observation is that while there are some benefits of considering smaller values for batch size, there are some chances that smaller batch sizes may yield overfitting.
\end{enumerate}

\vspace{-0.2cm}
\section{Limitations}
\label{sec:limitations}
The initial assumption of this work is that average users in many areas, including finance and economics are mostly interested in simple form of deep learning models including LSTM. Consequently, it may be relatively challenging for these users to build and fine tune deep learning models with complex architecture. To resemble this assumption, this paper also tries to keep the deep learning models simple without adding additional complexity to the architecture of the models built. In addition, it is also important to note that, to have a fair comparison between models generated by LLMs, it is important to avoid possible bias introduced by complexity of deep learning architecture. To prevent such unfair comparison, the work keeps the models at the very consistent and simple so the results can be justified without any bias.

Furthermore, different application domains may exhibit different results. This paper focuses only on financial data, as an important application domain. Including additional experiments and their results in some other application domains would make the paper very lengthy and confusing. Additional replication of the work performed here is necessary in different application domains.  According to our initial assumption, additional in-depth analysis might not be of interest for average researchers or developers with little background in this domain. As such, this paper focuses only on type of analysis that is often required by the average data analyst in some other application domains such as finance.

\vspace{-0.2cm}
\section{Conclusion and Future Work}
\label{sec:conclusion}
This paper reports the results of a number of controlled experiments to study the effect of various prompts with different sensitivity levels and configuration parameters of LLMs on the goodness of deep learning-based models generated for forecasting time series data. As a representative application domain, the paper studied the problem of forecasting financial time series data. The paper first created and optimized a manual LSTM-based model to forecast financial and stock time series data. We then controlled each prompt with respect to four criteria including 1) Clarity and Specificity, 2) Objective and Intent, 3) Contextual and Information, and 4) Format and Style where the sensitivity of these criteria were controlled in terms of being low, medium, and high. 

The results provided interesting insights regarding the accuracy of forecasting models generated by generative AI and LLMs. More notably, we observed that these generative AIs are capable to produce comparable forecasting models when queried using simple or complex prompts with additional details. We compared the accuracy of models with a single manually crafted and optimized LSTM-based forecasting model that was trained and built based on all datasets all together. According to our results, we did not observe significant influence of complex prompts to produce better and more accurate models. In some cases, the more simple prompts produced better and more accurate models; whereas, in some other cases, more complex prompts generated more accurate forecasting models. It is apparent that the value of temperature parameter used in configuring LLMs has direct impact on whether simple or more complex prompts can generate more accurate forecasting models. 

As for statistical performance, we observed that RMSE values for the models produced by LLMs are quite strong and the models remain robust. Additional statistical testing found differences between LLMs and manually coded models to be statistically significant, with particularly strong differences when datasets were more complex.

Moreover, we found that models generated by different LLMs used drastically different architectures, such as the number of layers, the number of units, and the activation functions. The differences in performance attributable to this variability may be an indication that prompt engineering is still an important feature for realizing LLMs for deep learning tasks.

The results reported in this paper are in particular useful for data analysts and practitioners who have little experience with programming and coding for developing complex deep learning-based models such as LSTM for forecasting time series data. The paper poses an interesting research problem that needs additional studies and expands the performance analysis to incorporate the metrics and statistical measures. Also, compare the LSTM model against models like ARIMA and other conventional models for further investigations to validate the results reported in this paper. 

\section{Acknowledgement}
This research is partially supported by the U.S. National Science Foundation Award: 2319802.

\bibliographystyle{cas-model2-names}

\bibliography{reference}





\end{document}